\PassOptionsToPackage{table}{xcolor}
\documentclass{style}

\usepackage{amsmath,amssymb}
\usepackage{url}
\usepackage{wrapfig}

\usepackage{amsmath,amsfonts,bm}
\usepackage{pifont}

\def\eqref#1{equation~\ref{#1}}

\def\1{\bm{1}}

\DeclareMathAlphabet{\mathsfit}{\encodingdefault}{\sfdefault}{m}{sl}
\SetMathAlphabet{\mathsfit}{bold}{\encodingdefault}{\sfdefault}{bx}{n}

\newcommand{\graytag}[1]{%
    \begingroup
    \setlength{\fboxsep}{1pt}%
    \smash{\raisebox{-0.05ex}{\colorbox{gray!15}{#1}}}%
    \endgroup
}

\title{SimForcing: Distilling Simulation Motion Priors into Real-Domain Robot World Models}
\author[1]{Xiaodong Wang}
\author[1]{Tianle Li}
\author[1]{Chuanxin Song}
\author[2]{Junliang Xie}
\author[2]{Zhanmi Zhong}
\author[2]{Suiying Wu}
\author[1\ddagger]{Peixi Peng}

\affiliation{$^1$Peking University.\ \   $^2$Guangzhou Frontop Digital Creative Technology Co., Ltd.}
\contribution[\ddagger]{Corresponding Author}

\abstract{
Action-conditioned robot world models must respond precisely to robot trajectories while preserving realistic visual dynamics, yet learning both from heterogeneous robot videos remains challenging. Simulation offers structured motion supervision, but appearance differences hinder direct transfer, and inaccurate simulation predictions can misguide real-video generation. We present SimForcing, a simulation-guided framework that uses simulation both as a source of transferable motion knowledge and as a controllable reference for prediction. First, we transfer motion knowledge from a simulation teacher through latent-motion distillation, aligning temporal changes in latent space to internalize motion priors while mitigating the influence of appearance differences. Second, we introduce multi-block simulation conditioning with condition dropout to exploit predicted simulation trajectories without relying excessively on their accuracy. Our simulation-conditioning classifier-free guidance scheme unifies these two ideas by balancing predictions based on internalized motion knowledge with those additionally guided by simulation latents. The jointly trained student generates both simulation conditions and real-domain videos, requiring no additional world model at inference. On Bridge, SimForcing achieves the best PSNR, SSIM, LPIPS, and FVD among the compared methods without external embodied pretraining. Evaluation on InternData-A1 further supports its applicability across robot datasets. Moreover, using our trained world model to initialize a vision-language-action model improves LIBERO success, suggesting its utility for downstream policy learning.

\vspace{1\baselineskip}
\textbf{GitHub}: \url{https://github.com/Wang-Xiaodong1899/SimForcing}
\vspace{-10mm}
}

\begin{document}
\maketitle

\begin{figure}[h]
\vspace{-6mm}
    \centering
    \includegraphics[width=0.95\linewidth]{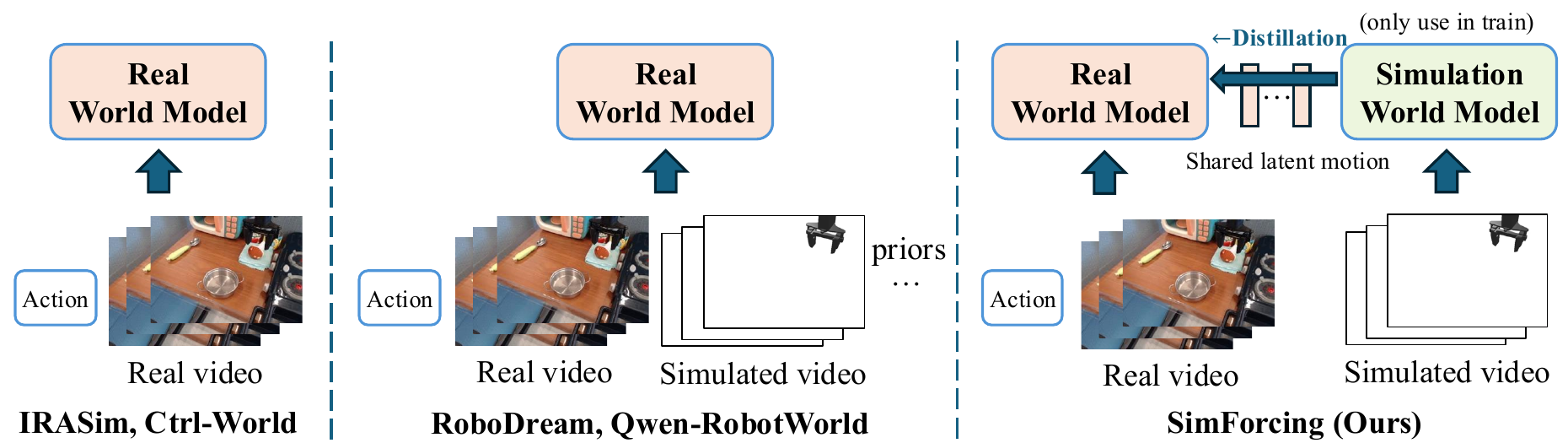}
    \vspace{-2mm}
\caption{\textbf{Comparison of training paradigms for action-conditioned robot world models.}
\textbf{Left:} Direct training on real videos and corresponding actions~\citep{zhu2024irasim,guo2025ctrlworld}.
\textbf{Middle:} Training on real videos augmented with explicit visual priors, such as simulated videos, scene priors, and object priors~\citep{zhang2026qwenrobotworld,ye2026robodream}.
\textbf{Right:} SimForcing learns a simulation world model from synthetic rollouts and distills its motion priors into a real-world model through latent-space motion supervision, without requiring additional scene or object decomposition.}
    \label{fig:teaser}
\end{figure}
\section{Introduction}
\label{sec:introduction}
Recent advances in large-scale video generation and world
foundation models~\citep{wan2025wan,chi2025wow,ali2025cosmos2.5,ma2026lingbotvideo},
have expanded the potential of learned visual dynamics
for embodied intelligence.
Beyond video synthesis, these models can generate training
experiences for robot policies~\citep{jang2025dreamgen}
and support the improvement of vision-language-action
(VLA) models through synthetic
rollouts~\citep{guo2026vlaw}.
Benchmarks such as WorldArena~\citep{shang2026worldarena}
further assess their utility as data engines, policy
evaluators, and action planners.
Despite this progress, visually convincing predictions
do not necessarily ensure reliable physical interactions
or strong downstream performance~\citep{shang2026worldarena}.
For robotic manipulation, a key challenge remains:
learning precise responses to continuous action trajectories
while preserving realistic appearance and temporally
coherent scene dynamics.

Existing approaches address this problem through different
training paradigms, as illustrated in Fig.~\ref{fig:teaser}.
One approach learns video prediction directly from real robot
videos and their associated actions such as
IRASim~\citep{zhu2024irasim} and
Ctrl-World~\citep{guo2025ctrlworld}.
This provides direct supervision for action-conditioned generation,
but learning precise action responses and complex visual dynamics
jointly from real videos remains challenging.
Another approach introduces an explicit visual structure.
For example, RoboDream~\citep{ye2026robodream} anchors generation
to rendered robot motion and conditions on scene and object
priors. Such compositional inputs provide additional control,
while requiring the corresponding priors to be constructed
and supplied. Qwen-RobotWorld~\citep{zhang2026qwenrobotworld} also needs scene video as prior. However, the potential of simulation data to improve real-video prediction without additional priors remains
underexplored.

Simulation provides structured supervision for action-dependent motion, complementing the appearance and interaction diversity of real videos. Transferring this knowledge nevertheless presents two challenges. First, appearance differences complicate direct matching between simulated and real observations. Second, predicted simulation trajectories are imperfect; a real-domain model that relies too heavily on them may inherit their errors. These challenges motivate transferring motion knowledge through temporal differences in latent space to mitigate the influence of appearance differences, while controlling the model's reliance on potentially inaccurate simulation predictions.

We present \textbf{SimForcing}, a novel world modeling framework that addresses these challenges through motion knowledge transfer and controllable simulation guidance. To transfer motion knowledge across domains with different appearances, we first train a mixture-of-transformers world model on synthetic rollouts with trajectory augmentation, learning action-dependent motion from diverse robot trajectories. The resulting model initializes the student and serves as a frozen teacher. Our key transfer mechanism is latent-motion distillation, which aligns temporal differences between adjacent teacher and student video latents rather than directly matching their appearances. This motion-focused supervision internalizes the teacher's simulation-derived priors within the student, directly
enhancing its ability to predict real-world video dynamics. Joint flow-matching and motion-distillation objectives on paired simulated and real videos further develop prediction capabilities in both domains within the same student, without requiring additional scene or object decomposition.

To benefit from explicit simulation predictions without becoming overly dependent on their accuracy, we introduce multi-block simulation conditioning with condition dropout. Injecting simulation latents into multiple video transformer blocks provides a motion reference throughout generation, while latent corruption and condition dropout train the student to handle imperfect or absent simulation guidance. At inference, we introduce a simulation-conditioning classifier-free guidance (CFG) that combines predictions from branches with and without simulation latent input, balancing explicit simulation guidance with the student's internalized motion priors. This design enables the model to exploit reliable simulation predictions while limiting the influence of inaccurate ones. The jointly trained student generates both the simulation conditions and real-domain videos, requiring no additional world model at inference.

Experiments on Bridge show that SimForcing outperforms
Ctrl-World, EnerVerse-AC, and a GeniWorld reimplementation
across PSNR, SSIM, LPIPS, and FVD, improving both reconstruction
fidelity and perceptual similarity.
It also achieves lower LPIPS than Cosmos-Predict2.5 with
substantially faster inference.
Evaluation on InternData-A1 supports its applicability across
robot datasets, while ablations demonstrate the complementary
benefits of motion distillation and simulation conditioning.
Beyond video prediction, initializing a downstream VLA with
the pretrained world-model experts improves average LIBERO
success from $82.6\%$ to $89.2\%$ under action-only training
relative to Wan DiT initialization, with gains also observed
in the other training settings. Our contributions are threefold:
\begin{itemize}
    \item We introduce SimForcing, a robot world model framework for transferring
    motion knowledge from simulation to real-world video
    prediction in latent space, internalizing simulation-derived
    motion priors within the world model.

    \item We propose a controllable mechanism for exploiting
    simulation predictions while limiting overreliance on
    inaccurate guidance. Our simulation-conditioning CFG scheme
    combines this explicit guidance with the motion knowledge
    internalized through distillation.

    \item We outperform state-of-the-art methods on Bridge across
    PSNR, SSIM, LPIPS, and FVD, and demonstrate the framework's
    applicability to InternData-A1 dataset.
    LIBERO experiments preliminarily validate the pretrained
    video expert's utility for policy learning.
\end{itemize}
\section{Related Work}
\label{sec:related_work}

\subsection{Generative World Models for Embodied Interaction}
Large-scale video generators provide increasingly capable visual
priors for world modeling~\citep{wan2025wan,agarwal2025cosmospredict1,ali2025cosmos2.5,chi2025wow},
with recent efforts extending pretraining and adaptation toward
embodied interaction.
Qwen-RobotWorld~\citep{zhang2026qwenrobotworld} uses
natural-language actions to predict future visual trajectories
across manipulation, navigation, and driving, while
LingBot-Video~\citep{ma2026lingbotvideo} scales
mixture-of-experts video pretraining with robot-oriented data.
DreamGen~\citep{jang2025dreamgen} adapts video generators to
target robot embodiments and recovers action labels through
inverse dynamics or latent action models, producing synthetic
trajectories for policy learning.
RoboScape~\citep{shang2025roboscape} incorporates physical
priors through joint RGB--depth prediction and keypoint-based
temporal consistency.
PhysisForcing~\citep{zhang2026physisforcing} improves physical
consistency using supervision constructed through a multi-model
pipeline involving point tracking, depth-aware region selection,
and semantic relation extraction.
In contrast, our latent-motion distillation derives supervision
directly from temporal differences in a simulation-pretrained
teacher's predicted video latents, without constructing explicit
depth, point-trajectory, or semantic-relation targets.

Action-conditioned world models connect robot controls to
predicted visual outcomes.
IRASim~\citep{zhu2024irasim} learns from historical observations
and continuous actions, while AVID~\citep{rigter2024avid} adds
an action-conditioned adapter to a frozen video diffusion model.
Dynamic World Simulation~\citep{he2025worldsim} combines action
conditioning with a motion-reinforced objective.
EnerVerse-AC~\citep{jiang2025enerverse} uses multi-level action
conditioning for multiview generation, and
Ctrl-World~\citep{guo2025ctrlworld} supports policy-in-the-loop
simulation with frame-level control and temporal memory.
Other methods encode control as spatial visual signals:
RoboMaster~\citep{fu2025robomaster} uses collaborative robot--object
trajectories, ABot-PhysWorld~\citep{chen2026abot} combines spatial
action injection with physics-aligned training, and
OSCAR~\citep{wu2026oscar} uses rendered kinematic skeletons
across embodiments.
EA-WM~\citep{yang2026eawm} projects actions and kinematic
states into camera-aligned visual action fields, with event-aware
bidirectional fusion supervised by VAE-encoded frame differences.
Masked Visual Actions~\citep{alzayer2026masked} reveals robot
or object motion in masked videos, unifying forward and inverse
world modeling through a shared pixel-space control interface.
FlowWAM~\citep{chen2026flowwam} adopts optical flow for video
prediction and policy learning, while
RynnWorld-Teleop~\citep{zhao2026rynnworld} uses hand-pose
conditioning for real-time teleoperation.

Recent efforts further incorporate 3D and 4D structure into
embodied prediction and control.
TesserAct~\citep{zhen2025tesseract} jointly generates RGB,
depth, and normal sequences for 4D scene reconstruction.
Robo4DGen~\citep{liu2025robo4dgen} introduces cross-view
pointmap supervision for geometrically consistent multiview
prediction, while MVISTA-4D~\citep{wang2026mvista} generates
arbitrary-view RGB-D futures through cross-view and
cross-modality feature fusion.
RoboStereo~\citep{zhang2026robostereo} couples action-conditioned
RGB and pointmap generation through a dual-tower diffusion
architecture, and RynnWorld-4D~\citep{zhao2026rynnworld4d}
jointly predicts RGB, depth, and optical flow to model
appearance, geometry, and motion.
Kinema4D~\citep{xu2026kinema4d} uses URDF-based kinematics
to construct robot pointmap trajectories as control signals
for synchronized RGB and pointmap generation.
CausalWM~\citep{xu2026causalwm} sequentially predicts optical
flow, 3D pointmaps, and RGB videos, using intermediate
predictions as context for subsequent generation.
PointWorld~\citep{huang2026pointworld} instead models scene
dynamics directly as 3D point flows, representing robot actions
in the same space and learning across embodiments from real
and simulated manipulation data.
X-WAM~\citep{guo2026xwam} further couples multiview RGB-D
prediction with robot action generation, using a lightweight
depth branch and asynchronous denoising to decode actions
with fewer steps than video.
WAM4D~\citep{li2026wam4d} uses spatial register tokens and
future-depth supervision to transfer geometric priors into
video--action representations, removing the geometric readout
branch for efficient action inference.
Our work addresses the complementary problem of transferring
simulation-derived motion knowledge into real-domain video
prediction, combining latent-motion distillation with
simulation guidance generated by the same model.

\subsection{World Action Models}
World action models couple visual prediction with robot control.
VPP~\citep{hu2024vpp}, Video Policy~\citep{liang2025videopolicy},
DiT4DiT~\citep{ma2026dit4dit}, and
mimic-video~\citep{pai2025mimicvideo} learn action policies from
predictive video representations.
Unified World Models~\citep{zhu2025uwm},
Motus~\citep{bi2025motus}, LingBot-VA~\citep{li2026lingbotva},
and DreamZero~\citep{ye2026dreamzero} jointly model future
observations and actions, connecting world modeling with
policy learning.
UniT~\citep{chen2026unit} develops shared visual-action tokens
for human-to-humanoid transfer, while
Cosmos Policy~\citep{kim2026cosmospolicy} and
World-Value-Action Model~\citep{li2026worldvalueaction}
incorporate value prediction for planning.
Future RGB modeling is a common visual learning objective in
video-based WAMs~\citep{zhu2025uwm,ye2026dreamzero}, although
auxiliary representations vary; Motus~\citep{bi2025motus},
for example, also learns latent actions from optical flow.

Beyond direct action generation, learned world models support
policy evaluation~\citep{li2025worldeval,li2026dworldeval},
reinforcement-learning-based policy improvement~\citep{jiang2026wovr},
and paired video-action data synthesis~\citep{lang2026vag}.
Kinema4D~\citep{xu2026kinema4d} and
RoboStereo~\citep{zhang2026robostereo} extend simulation with
explicit geometry, with RoboStereo further supporting policy
optimization.
Persistent Robot World Models~\citep{bardhan2026persistent}
and World Action Verifier~\citep{liu2026worldactionverifier}
improve model reliability through rollout-based reinforcement
learning and forward-inverse consistency, respectively.
Interactive World Simulator~\citep{wang2026interactive}
supports policy training and evaluation through learned
interactive dynamics, while
Genie Envisioner~\citep{liao2025genieenvisioner} and
Cosmos 3~\citep{nvidia2026cosmos3} integrate world modeling,
simulation, and control within broader frameworks.
FATE~\citep{wei2026fate} addresses the complementary problem
of feasibility-aware task generation and repair.

Recent work also revisits the visual supervision and generative
backbones used for control.
MaskWAM~\citep{yu2026maskwam} jointly predicts future RGB
observations, task-relevant masks, and actions, using optional
initial mask prompts to ground target selection.
ImageWAM~\citep{zhang2026imagewam} instead replaces dense
future-video modeling with a source-to-target-frame editing
objective and conditions an action expert on intermediate
editing features, without decoding the target image at inference.
Complementing these policy-oriented designs, our method focuses
on action-conditioned real-domain video prediction.
It distills simulation-derived motion through temporal latent
differences and uses the same model to generate simulation
guidance from input actions and a segmented robot observation,
without external robot-motion rendering at inference.

\subsection{Simulation Supervision and Real-domain Adaptation}
Simulation supports robot learning through data augmentation,
dynamics transfer, and visual conditioning.
Sim-and-Real Co-Training~\citep{maddukuri2025simreal} mixes
synthetic and real demonstrations for policy learning, while
RoboTwin 2.0~\citep{chen2025robotwin} scales synthetic data
generation with structured domain randomization.
ReDRAW~\citep{lanier2025redraw} adapts simulation-pretrained
latent dynamics through residual corrections, and Simulation
Distillation~\citep{levy2026simdist} transfers simulator priors
into world models for real-world adaptation and planning.
For video generation,
Cosmos-Transfer1~\citep{alhaija2025cosmos-transfer1} supports
sim-to-real visual translation through spatial controls.
GeniWorld~\citep{gu2026geniworld} and
AnchorDream~\citep{ye2025anchordream} condition video generation
on rendered robot motion, while
RoboDream~\citep{ye2026robodream} additionally incorporates
explicit scene and object priors.
These methods rely on external robot-motion rendering to
construct conditioning inputs at inference.
In contrast, our method uses the same model for both
simulation-domain and real-scene prediction, without invoking
an external simulator at inference.
Conditioned on the input actions and a robot-only image obtained
by semantically segmenting the initial observation, the model
predicts simulation-domain video latents that then guide its
real-scene video prediction.
This model-generated guidance complements the motion knowledge
internalized through latent-motion distillation.
Our simulation-conditioning CFG scheme controls its influence
to limit reliance on inaccurate predictions.
\begin{figure}[t]
    \centering
    \includegraphics[width=\linewidth]{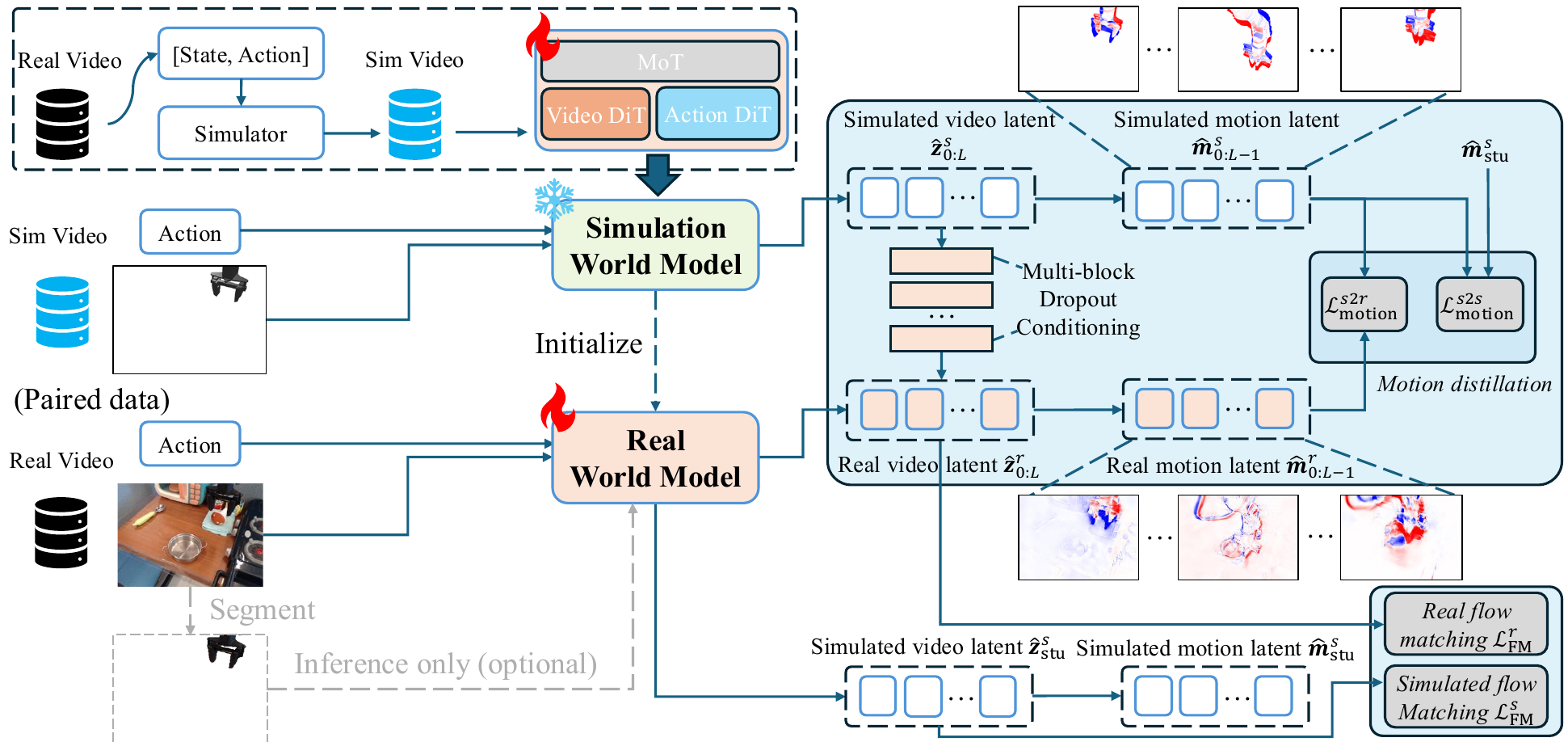}
    \caption{\textbf{Overview of SimForcing.}
\textbf{Simulation world model training:} A mixture-of-transformers world
model learns action-conditioned motion from augmented synthetic rollouts.
\textbf{Sim-to-real distillation:} The pretrained model initializes
a student and serves as a frozen teacher.
Latent-motion distillation internalizes motion priors, while
multi-block dropout conditioning provides simulation guidance.
Joint training enables the final student to generate both
simulation conditions and real videos, with CFG balancing
explicit guidance and internalized knowledge at inference.}
    \label{fig:train}
\end{figure}
\section{Method}
\label{sec:method}

\subsection{Problem formulation}
Let $\mathbf{x}_{0:T}=(\mathbf{x}_0,\ldots,\mathbf{x}_T)$ denote a robot video and let $\mathbf{a}_{0:T-1}$ denote its aligned low-level action trajectory. At each time step $t$, the proprioceptive state is synchronized with $\mathbf{x}_t$, and $\mathbf{a}_t$ is the command applied over the transition from $\mathbf{x}_t$ to $\mathbf{x}_{t+1}$. Each action comprises an end-effector translation increment, a rotation increment represented as an axis-angle vector, and a gripper command. The proprioceptive state comprises the end-effector position, orientation in Euler angles, and gripper state. Thus, a sequence of $T+1$ video frames is paired with $T$ actions and $T$ proprioceptive states.

Given the initial observation and future actions, our goal is to model
\begin{equation}
    p_\theta\!\left(\mathbf{x}_{1:T}\mid \mathbf{x}_0,\mathbf{a}_{0:T-1}\right).
    \label{eq:world_model}
\end{equation}
Proprioceptive inputs are used throughout but omitted from the notation for simplicity. The model predicts future visual observations; it does not predict actions during world-model pretraining. We use superscripts $s$ and $r$ for simulated and real samples.
A frozen video autoencoder $E$ compresses the video
$\mathbf{x}_{0:T}$ into a latent sequence
$\mathbf{z}_{0:L}=E(\mathbf{x}_{0:T})$,
where $\mathbf{z}_0$ represents the initial observation
and $L$ is the number of future latent time steps.
Due to temporal compression, $L<T$.

\subsection{Simulation World Model Training}
Learning motion responses and visual appearance jointly from
real videos is challenging.
As shown in Fig.~\ref{fig:train}, SimForcing first learns
action-conditioned robot motion in simulation, then transfers
this knowledge to real-video prediction through latent-motion
distillation and simulation conditioning.

The backbone combines a Video DiT and an Action DiT through
Mixture-of-Transformers (MoT) attention.
The Video DiT processes noisy future-video tokens and clean
initial-frame tokens, while the Action DiT encodes
$\mathbf{a}_{0:T-1}$ as conditioning tokens.
Joint attention connects the two streams, with modality-specific
feed-forward blocks processing their respective features.
Only the video branch receives a diffusion prediction loss.

We construct synthetic action--video pairs by replaying
trajectories from real demonstrations in a simulator.
Each rollout starts from the corresponding initial robot
configuration and renders motion aligned with the action sequence.

To increase motion diversity, we generate three perturbed variants
of each trajectory. We write each action as
$\mathbf{a}_t=(\mathbf{q}_t,g_t)$, where
$\mathbf{q}_t\in\mathbb{R}^6$ contains the translation and axis-angle
rotation increments, and $g_t$ is the gripper command.
The perturbed action is
\begin{equation}
\widetilde{\mathbf{a}}_t
=
\left(
\operatorname{clip}
\left(
(\mathbf{q}_t+\mathbf{b})\odot\mathbf{s}
+\boldsymbol{\eta}_t
\right),
g_t
\right),
\qquad
\boldsymbol{\eta}_t\sim\mathcal{N}(\mathbf{0},\boldsymbol{\Sigma}),
\label{eq:action_augmentation}
\end{equation}
where the bias $\mathbf{b}$ and element-wise scale $\mathbf{s}$
are uniformly sampled once per rollout, while
$\boldsymbol{\eta}_t$ is sampled independently at each step.
Each perturbed trajectory is replayed from the same initial robot
configuration to render a video paired with the executed actions.
For simplicity, we subsequently use $\mathbf{a}_{0:T-1}$ to denote
the executed action sequence, including perturbed variants.

Given a synthetic rollout consisting of $T+1$ video frames $\mathbf{x}^s_{0:T}$ and $T$ aligned actions $\mathbf{a}_{0:T-1}$,
let $\mathbf{z}_0^s$ denote the initial observation latent
and $\mathbf{z}^s=\mathbf{z}_{1:L}^s$ denote the clean
future latent sequence.
For Gaussian noise $\boldsymbol\epsilon\sim\mathcal{N}(\mathbf{0},\mathbf{I})$ and flow time $\tau\in[0,1]$, we define the interpolation path and target velocity as
\begin{equation}
    \mathbf{z}^s(\tau)=(1-\tau)\mathbf{z}^s+\tau\boldsymbol\epsilon,
    \qquad
    \mathbf{u}^s=\boldsymbol\epsilon-\mathbf{z}^s.
    \label{eq:flow_path}
\end{equation}
The simulation world model $S_\theta$ predicts the target velocity conditioned on the initial observation latent and action tokens, with proprioceptive inputs omitted from the notation as described above. We train the entire video-action backbone using the flow-matching objective
\begin{equation}
    \mathcal{L}_{\mathrm{vid}}^{s}
    =\mathbb{E}\left[
      \left\|
      S_\theta\!\left(
      \mathbf{z}^{s}(\tau),\tau\mid
      \mathbf{z}^{s}_0,\mathbf{a}_{0:T-1}
      \right)-\mathbf{u}^s
      \right\|_2^2
    \right].
    \label{eq:video_loss}
\end{equation}
The expectation is taken over synthetic rollouts, Gaussian noise, and sampled flow times.

The resulting simulation world model serves two purposes: it provides an initialization for action-conditioned video generation in the real domain and serves as a frozen teacher that supplies dynamics supervision during real-domain adaptation.

\subsection{Sim-to-Real Distillation Training}
The core idea of this part is to transfer motion priors from simulation to the real domain through latent-space distillation. During simulation pretraining, the world model learns to predict future robot motion from the initial observation and action sequence. By excluding object interactions and background complexity, this training focuses on the robot's action-conditioned motion dynamics. The learned latent representations then provide motion supervision and conditioning signals for training the real-domain world model. 



\paragraph{Motion Distillation.}
After simulation pretraining, we freeze the teacher parameters
$\bar{\theta}$ and initialize the student's video--action
backbone from the teacher.
The student parameters $\theta$ are optimized during subsequent
training, while $\bar{\theta}$ remains fixed.
The newly introduced conditioning projections are initialized
to zero.
We represent latent motion as the difference between temporally adjacent video latents:
\begin{equation}
    \mathbf{m}_{\ell}
    = \mathbf{z}_{\ell+1}-\mathbf{z}_{\ell},
    \qquad \ell=0,\ldots,L-1.
    \label{eq:latent_motion}
\end{equation}
Here, $\ell$ indexes the temporally compressed latent sequence,
and $\mathbf{z}_0$ is the initial observation latent.

To obtain motion supervision on the fly during training, we follow the flow-matching training procedure, using a single noise injection and velocity prediction for each domain. Specifically, we construct the noisy latents using the same Gaussian noise $\boldsymbol\epsilon$ and flow time $\tau$:
\begin{equation}
    \mathbf{z}^d(\tau)
    = (1-\tau)\mathbf{z}^d+\tau\boldsymbol\epsilon,
    \qquad d\in\{s,r\}.
    \label{eq:paired_noise}
\end{equation}
We then perform a single forward pass through the frozen simulation teacher $S_{\bar{\theta}}$ and the trainable real-domain student $R_\theta$ to predict their respective velocities:
\begin{equation}
    \begin{aligned}
    \hat{\mathbf{u}}^s
    &= S_{\bar{\theta}}\!\left(
        \mathbf{z}^s(\tau),\tau
        \mid \mathbf{z}_0^s,\mathbf{a}_{0:T-1}
    \right),\\
    \hat{\mathbf{u}}^r
    &= R_\theta\!\left(
        \mathbf{z}^r(\tau),\tau
        \mid \mathbf{z}_0^r,\mathbf{a}_{0:T-1};
        \mathbf{c}^s,g
    \right).
    \end{aligned}
    \label{eq:paired_velocity}
\end{equation}
Here, $\mathbf{c}^s$ and $g$ denote the simulation condition
and its dropout gate, defined below.

Following the interpolation path in Eq.~\ref{eq:flow_path}, we estimate the clean future latent from each predicted velocity:
\begin{equation}
    \hat{\mathbf{z}}^d
    = \mathbf{z}^d(\tau)-\tau\hat{\mathbf{u}}^d,
    \qquad d\in\{s,r\}.
    \label{eq:clean_latent_estimate}
\end{equation}
We compute the predicted latent motion as the difference between adjacent latents:
\begin{equation}
    \hat{\mathbf{m}}_{\ell}^d
    = \hat{\mathbf{z}}_{\ell+1}^d
      -\hat{\mathbf{z}}_{\ell}^d,
    \qquad \ell=0,\ldots,L-1,\quad d\in\{s,r\}.
    \label{eq:predicted_latent_motion}
\end{equation}
Here, $\hat{\mathbf{z}}_0^d=\mathbf{z}_0^d$ is the clean
initial observation latent, and
$\hat{\mathbf{m}}^d
=(\hat{\mathbf{m}}_0^d,\ldots,\hat{\mathbf{m}}_{L-1}^d)$
denotes the predicted motion sequence. We then minimize the squared $L_2$ distance between the student's and teacher's latent motion sequences:
\begin{equation}
    \mathcal{L}^{s2r}_{\mathrm{motion}}
    = \mathbb{E}\left[
        \left\|\hat{\mathbf{m}}^r-\hat{\mathbf{m}}^s\right\|_2^2
    \right].
    \label{eq:motion_distillation}
\end{equation}
The simulation teacher remains frozen, and gradients are propagated only through the real-domain student. This objective constrains the predicted motion while encouraging temporal consistency in the background, helping suppress background flickering as illustrated in Fig.~\ref{fig:train}.

\paragraph{Multi-block dropout conditioning.}
To complement the motion priors distilled into the student, we introduce multi-block dropout conditioning, which provides simulation latents as an optional source of motion guidance. Specifically, we inject the simulation latent into every video transformer block through an independent, zero-initialized 3D convolutional projection. Each projection maps the simulation latent to features aligned with the video tokens, which are added before the feed-forward network:
\begin{equation}
    \mathbf{h}_j \leftarrow
    \mathbf{h}_j + g\,P_j(\mathbf{c}^s),
    \qquad
    g\sim\operatorname{Bernoulli}(p),
    \label{eq:multiblock_conditioning}
\end{equation}
where $\mathbf{h}_j$ denotes the video tokens in block $j$, $P_j$ is its condition projection, and $\mathbf{c}^s$ is the simulation latent condition. The gate $g$ is sampled independently for each training sample and shared across all blocks, with $p$ denoting the condition retention probability. This design supplies motion information throughout the network. Zero initialization preserves the pretrained backbone's behavior at the start of training, allowing the conditioning pathways to be learned gradually.

To improve robustness to imperfect simulation predictions, we stochastically corrupt the ground-truth simulation latent during training:
\begin{equation}
    \mathbf{c}^s
    = \alpha\mathbf{z}^s
      +(1-\alpha)\sigma_c\boldsymbol{\eta},
    \qquad
    \boldsymbol{\eta}\sim\mathcal{N}(\mathbf{0},\mathbf{I}),
    \label{eq:condition_corruption}
\end{equation}
where $\alpha$ and $\sigma_c$ are sampled per example, and $\boldsymbol{\eta}$ is independent of the flow-matching noise. This corruption is applied with a fixed probability; otherwise, we use the clean simulation latent. It affects only the conditioning input, leaving the flow-matching and teacher supervision targets unchanged.

Condition dropout prevents the student from relying excessively on simulation latents. When $g=0$, all simulation-conditioning residuals are disabled, requiring the student to predict future observations using the motion knowledge distilled into its own parameters. When $g=1$, the student additionally receives simulation guidance at every video block. Both paths retain the initial observation, actions, and motion distillation remains active for all samples regardless of the gate.

To further exploit paired data and consolidate prediction in both domains within a single model, we train the student on both simulation and real videos. The simulation branch uses the simulation initial observation and actions without simulation latent conditioning, while the real branch uses the dropout conditioning described above. Both branches share the same student parameters.

For joint training, let $\hat{\mathbf{u}}_{\mathrm{stu}}^d$
denote the student's velocity prediction in domain $d\in\{s,r\}$.
The real-domain prediction is
$\hat{\mathbf{u}}_{\mathrm{stu}}^r=\hat{\mathbf{u}}^r$
from Eq.~\ref{eq:paired_velocity}, while the simulation-domain
prediction disables simulation conditioning:
\begin{equation}
    \hat{\mathbf{u}}_{\mathrm{stu}}^s
    = R_\theta\!\left(
        \mathbf{z}^s(\tau),\tau
        \mid \mathbf{z}_0^s,\mathbf{a}_{0:T-1};
        g=0
    \right).
    \label{eq:student_sim_velocity}
\end{equation}
We apply flow matching in both domains:
\begin{equation}
    \mathcal{L}_{\mathrm{FM}}^d
    = \mathbb{E}\left[
        \left\|
        \hat{\mathbf{u}}_{\mathrm{stu}}^d
        -(\boldsymbol{\epsilon}-\mathbf{z}^d)
        \right\|_2^2
    \right],
    \qquad d\in\{s,r\}.
    \label{eq:joint_flow_matching}
\end{equation}
We also estimate the student's clean simulation latents as
$\hat{\mathbf{z}}_{\mathrm{stu}}^s
=\mathbf{z}^s(\tau)-\tau\hat{\mathbf{u}}_{\mathrm{stu}}^s$.
Using $\hat{\mathbf{z}}_{\mathrm{stu},0}^s=\mathbf{z}_0^s$,
we compute the motion sequence $\hat{\mathbf{m}}_{\mathrm{stu}}^s$
by adjacent differences and align it with the frozen teacher:
\begin{equation}
    \mathcal{L}_{\mathrm{motion}}^{s2s}
    = \mathbb{E}\left[
        \left\|
        \hat{\mathbf{m}}_{\mathrm{stu}}^s
        -\hat{\mathbf{m}}^s
        \right\|_2^2
    \right].
    \label{eq:simulation_motion_distillation}
\end{equation}
Here, $\hat{\mathbf{m}}^s$ is the frozen teacher's motion target.
Together with real-domain motion distillation, this objective
helps preserve the student's simulation prediction capability
during joint training.

\begin{table}[t]
    \centering
    \setlength{\belowcaptionskip}{6pt}
    \setlength{\tabcolsep}{3pt}
    \caption{Action-conditioned video prediction on the
    Bridge~\citep{walke2023bridgedata} validation set.
    \graytag{Gray} rows denote embodied pretrained models. Time denotes inference time per sample. Best results among models without embodied pretraining
    are in bold.}
    \label{tab:bridge}
    \begin{tabular}{@{}l|cccccc@{}}
        \toprule
        Model & Time (s) $\downarrow$
        & PSNR $\uparrow$ & SSIM $\uparrow$
        & FID $\downarrow$ & FVD $\downarrow$
        & LPIPS $\downarrow$ \\
        \midrule
        \multicolumn{7}{@{}l}{\textit{With embodied pretraining}} \\
        \rowcolor{gray!15}
        IRASim~\citep{zhu2024irasim}
        & 25.9 & 18.062 & 0.753 & 47.27 & 333.92 & 0.2102 \\
        \rowcolor{gray!15}
        Cosmos-Predict2.5~\citep{ali2025cosmos2.5}
        & 20.5 & 27.216 & 0.880 & 23.74 & 119.38 & 0.0996 \\
        \midrule
        \multicolumn{7}{@{}l}{\textit{Without embodied pretraining}} \\
        Ctrl-World~\citep{guo2025ctrlworld}
        & 16.9 & 20.928 & 0.761 & 46.67 & 235.57 & 0.1674 \\
        EnerVerse-AC~\citep{jiang2025enerverse}
        & 16.6 & 22.599 & 0.797 & 28.67 & 181.55 & 0.0933 \\
        Wan2.2-TI2V-5B (SFT)~\citep{wan2025wan}
        & \textbf{1.8} & 22.751 & 0.838 & 23.97 & 168.71 & 0.0926 \\
        GeniWorld~\citep{gu2026geniworld}
        & 4.1 & 23.436 & 0.845 & \textbf{23.62} & 148.74 & 0.0799 \\
        SimForcing (Ours)
        & 4.1 & \textbf{25.077} & \textbf{0.858}
        & 25.21 & \textbf{137.49} & \textbf{0.0673} \\
        \bottomrule
    \end{tabular}
\end{table}

The complete training objective comprises four terms:
\begin{equation}
    \mathcal{L}
    = \lambda_{\mathrm{FM}}^r\mathcal{L}_{\mathrm{FM}}^r
      +\lambda_{\mathrm{FM}}^s\mathcal{L}_{\mathrm{FM}}^s
      +\lambda_{\mathrm{motion}}^{s2r}\mathcal{L}_{\mathrm{motion}}^{s2r}
      +\lambda_{\mathrm{motion}}^{s2s}\mathcal{L}_{\mathrm{motion}}^{s2s}.
    \label{eq:total_loss}
\end{equation}
The teacher is used only during training. At inference, the student can generate real-domain videos using the following guidance method, eliminating the need for an additional simulation world model.

\begin{figure}
    \centering
    \includegraphics[width=\linewidth]{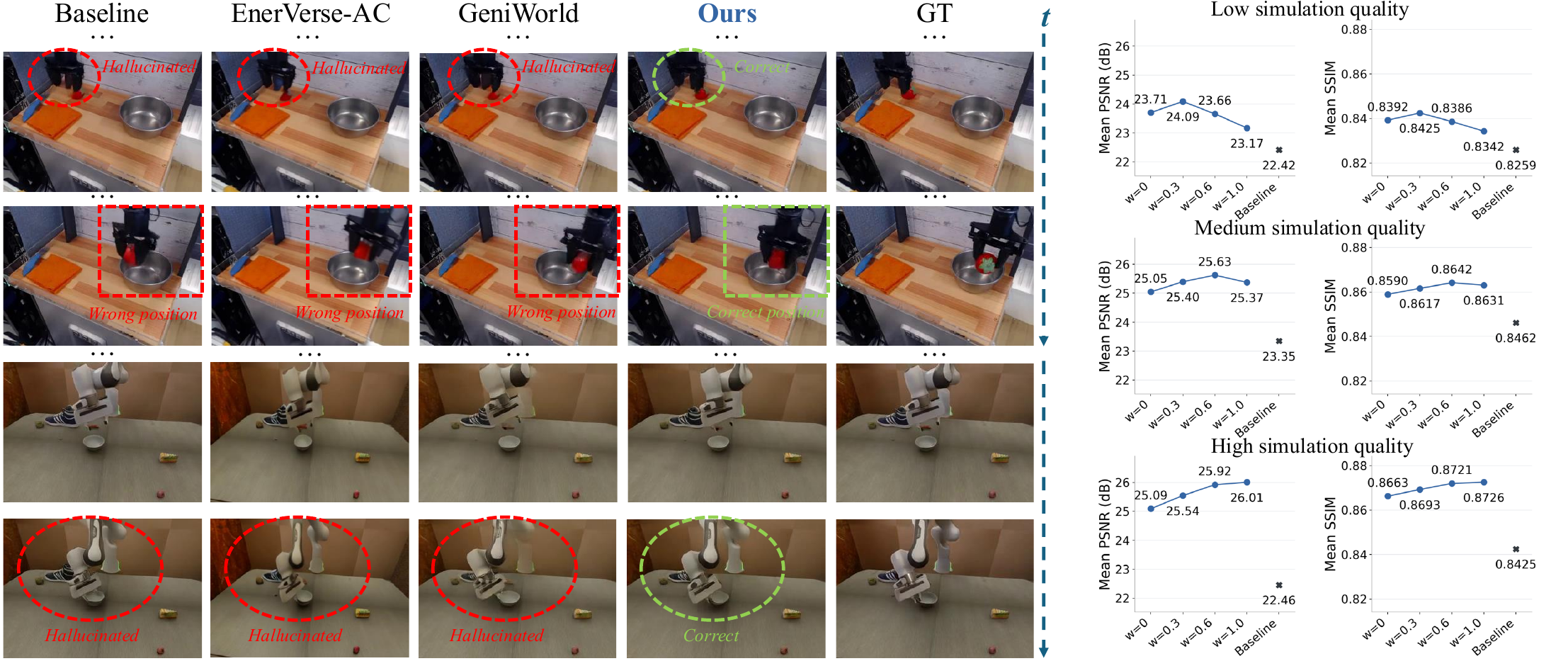}
    \vspace{-2mm}
    \caption{\textbf{Left:} Qualitative comparison across methods.
The first two rows show examples from Bridge~\citep{walke2023bridgedata},
and the last two from InternData-A1~\citep{tian2026interndata}.
Our method produces fewer visual hallucinations and more
accurate action-conditioned motion.
\textbf{Right:} Effect of simulation prediction quality
on our method's real-video predictions on Bridge.}
    \label{fig:vis1}
    \vspace{-4mm}
\end{figure}

\paragraph{Simulation-conditioning classifier-free guidance.}
At inference, we use SAM3~\citep{carion2026sam3} to segment
the robot arm in the initial real observation and encode
the masked observation as $\mathbf{z}_0^s$.
Conditioned on $\mathbf{z}_0^s$ and the action trajectory,
the student $R_\theta$ first generates simulation latents
with simulation conditioning disabled.
These latents form the condition $\mathbf{c}^s$, which remains
fixed throughout real-video sampling.
We then evaluate the same student with the conditioning
pathways enabled and disabled:
\begin{equation}
    \begin{aligned}
    \hat{\mathbf{u}}_{\mathrm{cond}}
    &= R_\theta\!\left(
        \mathbf{z}^r(\tau),\tau
        \mid \mathbf{z}_0^r,\mathbf{a}_{0:T-1};
        \mathbf{c}^s,g=1
    \right),\\
    \hat{\mathbf{u}}_{\mathrm{uncond}}
    &= R_\theta\!\left(
        \mathbf{z}^r(\tau),\tau
        \mid \mathbf{z}_0^r,\mathbf{a}_{0:T-1};
        g=0
    \right).
    \end{aligned}
    \label{eq:condition_cfg_branches}
\end{equation}
Here, $\mathbf{z}^r(\tau)$ denotes the current real-video
latent state during sampling.
Both branches retain the initial observation and action trajectory.
The guided velocity is
\begin{equation}
    \hat{\mathbf{u}}_{\mathrm{cfg}}
    = \hat{\mathbf{u}}_{\mathrm{uncond}}
      + w\left(
        \hat{\mathbf{u}}_{\mathrm{cond}}
        -\hat{\mathbf{u}}_{\mathrm{uncond}}
      \right).
    \label{eq:condition_cfg}
\end{equation}
Setting $w=0$ uses the student's internalized motion priors
without simulation conditioning, while $w=1$ recovers the
conditional prediction; intermediate weights balance the two.
The entire process requires only the final student model.
Although simulation guidance depends on the quality of
the student's own rollouts, $w$ controls reliance on them
at inference.
Fig.~\ref{fig:vis1} (right) shows consistent improvements
over the baseline across the displayed guidance weights.

\begin{table}[h]
    \centering
    \setlength{\belowcaptionskip}{6pt}
    \caption{Component ablations on Bridge. FM, TA, MD, and SC denote joint flow matching, trajectory augmentation, motion distillation,
    and simulation conditioning, respectively. Best are in bold.}
    \label{tab:ablation}
    \begin{tabular}{ccccc|ccccc}
        \toprule
        & \multicolumn{4}{c|}{Training components}
        & \multicolumn{5}{c}{Evaluation metrics} \\
        \cmidrule(lr){2-5} \cmidrule(lr){6-10}
        ID & FM & TA & MD & SC
        & PSNR $\uparrow$ & SSIM $\uparrow$
        & FID $\downarrow$ & FVD $\downarrow$ & LPIPS $\downarrow$ \\
        \midrule
        \multicolumn{5}{l|}{Wan2.2 (SFT)~\citep{wan2025wan}}
        & 22.751 & 0.838 & 23.97 & 168.71 & 0.0926 \\
        \midrule
        (a) & \ding{51} & \ding{55} & \ding{55} & \ding{55}
        & 24.128 & 0.851 & 22.03 & 141.81 & 0.0751 \\
        (b) & \ding{51} & \ding{51} & \ding{55} & \ding{55}
        & 24.523 & 0.854 & 21.52 & 134.56 & 0.0702 \\
        (c) & \ding{51} & \ding{55} & \ding{51} & \ding{55}
        & 24.166 & 0.849 & 27.46 & 154.69 & 0.0768 \\
        (d) & \ding{51} & \ding{51} & \ding{51} & \ding{55}
        & 24.310 & 0.852 & \textbf{19.18}
        & \textbf{100.27} & 0.0769 \\
        (e) & \ding{51} & \ding{51} & \ding{55} & \ding{51}
        & 23.732 & 0.848 & 21.83 & 145.40 & 0.0772 \\
        \midrule
        (f) & \ding{51} & \ding{51} & \ding{51} & \ding{51}
        & \textbf{25.077} & \textbf{0.858}
        & 25.21 & 137.49 & \textbf{0.0673} \\
        \bottomrule
    \end{tabular}
\end{table}

\section{Experiments}
\label{sec:experiments}

\subsection{Experimental Setup}
\label{sec:experimental_setup}
We conduct our main evaluation and ablations on
Bridge~\citep{walke2023bridgedata}, assess generalization on
InternData-A1~\citep{tian2026interndata}, and evaluate downstream
VLA learning on LIBERO~\citep{liu2023libero}.
For video prediction, models generate future frames from an
initial observation and an action trajectory.
The baseline combines Wan2.2-TI2V-5B~\citep{wan2025wan} with
a 1B Action DiT and is trained on real videos.
We compare with Cosmos-Predict2.5~\citep{ali2025cosmos2.5}, GeniWorld~\citep{gu2026geniworld},
IRASim~\citep{zhu2024irasim}, Ctrl-World~\citep{guo2025ctrlworld},
and EnerVerse-AC~\citep{jiang2025enerverse}.
SimForcing learns motion priors from synthetic rollouts and
trains a student on paired simulated and real videos under
a frozen simulation teacher. We report PSNR and SSIM for reconstruction fidelity,
LPIPS for perceptual similarity, and FID and FVD for image-
and video-level distributional quality.
Higher PSNR and SSIM and lower LPIPS, FID, and FVD are better.
For downstream policy learning, we report task success rates on LIBERO. More details can be found in the Appendix~\ref{app:in}.

\begin{table*}[t]
    \centering
    \caption{Ablation studies on distillation targets and CFG weights
    on the Bridge validation set. Bold indicates the best result
    within each comparison.}
    \label{tab:distillation_cfg}
    \small
    \setlength{\tabcolsep}{4pt}
    \renewcommand{\arraystretch}{1.15}
    \begin{tabular}{
    @{}l|ccccc
    @{\hspace{1.8em}}
    c|ccccc@{}
}
        \multicolumn{6}{c}{(a) Distillation target}
        & \multicolumn{6}{c}{(b) CFG weight} \\
        \cmidrule[\heavyrulewidth](r{0.8em}){1-6}
\cmidrule[\heavyrulewidth]{7-12}

        Target
        & PSNR $\uparrow$ & SSIM $\uparrow$
        & FID $\downarrow$ & FVD $\downarrow$ & LPIPS $\downarrow$
        & $w$
        & PSNR $\uparrow$ & SSIM $\uparrow$
        & FID $\downarrow$ & FVD $\downarrow$ & LPIPS $\downarrow$ \\
        \cmidrule(r{0.8em}){1-6}
        \cmidrule{7-12}

        \multirow{2}{*}{\shortstack[l]{Latent\\value}}
        & \multirow{2}{*}{19.232}
        & \multirow{2}{*}{0.803}
        & \multirow{2}{*}{\textbf{26.38}}
        & \multirow{2}{*}{193.15}
        & \multirow{2}{*}{0.0972}
        & 0.0 & 24.620 & 0.8549 & 25.08 & 145.96 & 0.0716 \\

        & & & & &
        & 0.3 & 25.014 & 0.8578 & 26.68 & 144.19 & 0.0687 \\

        \multirow{2}{*}{\shortstack[l]{Latent\\motion}}
        & \multirow{2}{*}{\textbf{24.166}}
        & \multirow{2}{*}{\textbf{0.849}}
        & \multirow{2}{*}{27.46}
        & \multirow{2}{*}{\textbf{154.69}}
        & \multirow{2}{*}{\textbf{0.0768}}
        & 0.6 & \textbf{25.077} & \textbf{0.8583}
        & 25.21 & 137.49 & \textbf{0.0673} \\

        & & & & &
        & 1.0 & 24.857 & 0.8567
        & \textbf{24.70} & \textbf{133.66} & 0.0681 \\

        \cmidrule[\heavyrulewidth](r{0.8em}){1-6}
\cmidrule[\heavyrulewidth]{7-12}
    \end{tabular}
\end{table*}

\begin{figure}[h]
    \centering
    \includegraphics[width=0.85\linewidth]{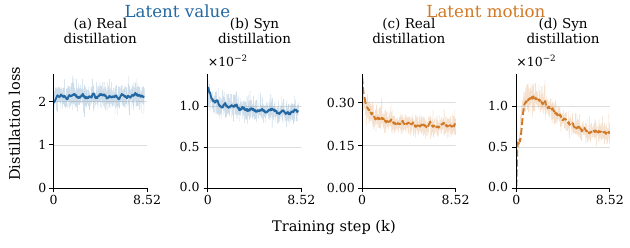}
    \caption{
        Distillation loss curves for latent value (a--b) and
        latent motion (c--d) on the real and synthetic branches.
        Faint curves show logged losses, and bold curves show
        trailing averages over 250 training steps.
        The real-branch panels use different vertical scales;
        the synthetic-branch panels share the same scale.
    }
    \label{fig:latent-distill-losses}
\end{figure}

\subsection{Video Prediction Evaluation}
\label{sec:bridge_results}
Tab.~\ref{tab:bridge} compares action-conditioned video prediction
on Bridge. Among models without embodied pretraining, SimForcing
achieves the best PSNR, SSIM, LPIPS, and FVD.
Compared with GeniWorld, it provides closer agreement with
action-conditioned reference videos at the same reported inference
time, with a slightly higher FID reflecting a trade-off between
prediction fidelity and image-level distributional quality.
Both methods perform segmentation at inference, requiring only
0.3~s per sample.
Cosmos-Predict2.5 achieves higher PSNR and SSIM and lower FVD,
while SimForcing attains lower LPIPS with approximately
five times faster inference.
These results support the benefits of simulation-derived priors
for prediction while maintaining practical inference costs. The qualitative comparisons in Fig.~\ref{fig:vis1} (left)
further illustrate these benefits on both Bridge and InternData-A1.
In the shown examples, compared to other methods, SimForcing produces fewer visual
hallucinations and motion that more closely follows the specified
actions.

\subsection{Analysis}
\label{sec:component_ablations}

\paragraph{Component ablations.}
Tab.~\ref{tab:ablation} shows that joint flow matching and
trajectory augmentation improve all five metrics over real-video
SFT. With augmentation, motion distillation achieves the lowest
FID and FVD, while combining it with simulation conditioning
yields the best PSNR, SSIM, and LPIPS.
This reflects a trade-off: the full model prioritizes prediction
fidelity and perceptual similarity to reference videos over
distributional quality.
Tab.~\ref{tab:distillation_cfg} further shows that latent-motion
distillation outperforms latent-value matching on four metrics,
supporting temporal differences as a transfer target that
mitigates the influence of cross-domain appearance differences.
CFG provides additional control over this trade-off:
$w=0.6$ achieves the best PSNR, SSIM, and LPIPS, whereas $w=1$
yields the lowest FID and FVD.
Fig.~\ref{fig:vis1} (right) further illustrates adjustable
simulation guidance, with improvements over the baseline
across the displayed weights.

\paragraph{Latent Value vs.\ Latent Motion Distillation.}
We compare two distillation objectives: latent value distillation,
which matches the student's predicted future latents to those of
the frozen teacher, and latent motion distillation, which instead
matches their differences between consecutive latent frames.
Fig.~\ref{fig:latent-distill-losses} presents the corresponding
distillation losses for the real and synthetic branches.
On the real branch, the latent value loss remains approximately
constant, whereas the latent motion loss decreases during training.
On the synthetic branch, both objectives decrease in the later
stages, with latent motion exhibiting an initial rise followed
by a sustained decline.
Since the two objectives supervise different quantities, their
absolute loss magnitudes should not be interpreted as a direct
comparison of model quality.

\begin{table}[]
    \centering
    \setlength{\belowcaptionskip}{6pt}
    \setlength{\tabcolsep}{3.5pt}
    \caption{Video prediction and component ablations on
InternData-A1. Best results are in bold.
FM, MD, SC denote joint flow matching, motion
distillation, and simulation conditioning, respectively.}
    \label{tab:generalize}
    \begin{tabular}{lccc|ccccc}
        \toprule
        Model & FM & MD & SC
        & PSNR $\uparrow$ & SSIM $\uparrow$
        & FID $\downarrow$ & FVD $\downarrow$ & LPIPS $\downarrow$ \\
        \midrule
        EnerVerse-AC~\citep{jiang2025enerverse}
        & -- & -- & --
        & 23.780 & 0.782 & 48.29 & 378.88 & 0.1056 \\
        Ctrl-World~\citep{guo2025ctrlworld}
        & -- & -- & --
        & 24.063 & 0.786 & 44.42 & 363.15 & 0.1099 \\
        Wan2.2 (SFT)~\citep{wan2025wan}
        & -- & -- & --
        & 24.738 & 0.808 & 30.94 & 335.20 & 0.0827 \\
        GeniWorld~\citep{gu2026geniworld}
        & -- & -- & --
        & 24.234 & 0.804 & 41.54 & 355.52 & 0.0962 \\
        \midrule
        \multicolumn{9}{l}{\textit{SimForcing component ablations}} \\
        Joint FM only
        & \ding{51} & \ding{55} & \ding{55}
        & 25.382 & \textbf{0.813} & \textbf{30.30}
        & 317.05 & \textbf{0.0742} \\
        With motion distillation
        & \ding{51} & \ding{51} & \ding{55}
        & 25.428 & \textbf{0.813} & 31.99 & 327.89 & 0.0761 \\
        With simulation conditioning
        & \ding{51} & \ding{55} & \ding{51}
        & 25.029 & 0.808 & 32.06 & 318.78 & 0.0806 \\
        SimForcing (full)
        & \ding{51} & \ding{51} & \ding{51}
        & \textbf{25.562} & \textbf{0.813}
        & 32.51 & \textbf{314.21} & 0.0765 \\
        \bottomrule
    \end{tabular}
\end{table}

\begin{table}[h]
    \centering
    \caption{LIBERO success rates (\%).
    Arrows indicate improvements over Wan2.2
    within each training setting.}
    \label{tab:libero}
    \begin{tabular}{@{}lccccc@{}}
        \toprule
        Init. & Spatial & Object & Goal & Long & Avg. \\
        \midrule
        FastWAM
        & 97.8 & 99.2 & 97.8 & 95.6 & 97.6 \\
        \midrule
        \multicolumn{6}{@{}l}{\textit{Action-only training}} \\
        Wan2.2
        & 88.0 & 95.6 & 82.8 & 64.0 & 82.6 \\
        Our WM
        & 89.4 & 95.6 & 92.4 & 79.2 & 89.2$\uparrow$ \\
        \midrule
        \multicolumn{6}{@{}l}{\textit{Unconditional training}} \\
        Wan2.2
        & 96.4 & 99.6 & 97.6 & 92.2 & 96.5 \\
        Our WM
        & 97.2 & 99.8 & 98.0 & 94.8 & 97.5$\uparrow$ \\
        \midrule
        \multicolumn{6}{@{}l}{\textit{Video + action training}} \\
        Wan2.2
        & 99.4 & 98.2 & 98.8 & 95.8 & 98.1 \\
        Our WM
        & 98.6 & 99.8 & 97.8 & 97.4 & 98.4$\uparrow$ \\
        \bottomrule
    \end{tabular}
\end{table}

\paragraph{Evaluation on InternData-A1.}
\label{sec:generalization}
Tab.~\ref{tab:generalize} evaluates our framework on
InternData-A1, with models trained on its training split.
The full model outperforms Ctrl-World, GeniWorld, and
EnerVerse-AC across all five metrics and improves over
Wan2.2 SFT on all metrics except FID.
It achieves the highest PSNR and lowest FVD, and ties for
the highest SSIM, while joint flow matching alone yields
the lowest FID and LPIPS.
These results support the framework's applicability to
another robot dataset, with different configurations
offering distinct benefits across metrics.

\paragraph{Downstream policy learning.}
Action-conditioned video prediction and downstream VLA learning
share the same model architecture, with all pretrained parameters
transferred to initialize the downstream model.
The action expert changes from a conditioning branch to an
action-prediction branch.
Tab.~\ref{tab:libero} shows that this initialization improves
average LIBERO success over Wan2.2 across all three
policy-training settings.
The largest gain occurs with action-only training, increasing
average success from $82.6\%$ to $89.2\%$, particularly through
improvements on Goal and Long.
Smaller gains persist with unconditional and joint video--action
training, with consistent improvements on Long across settings.
These results provide preliminary evidence that the pretrained
world-model representations benefit downstream policy learning.

\vspace{-2mm}
\section{Conclusion}
\vspace{-2mm}
We present SimForcing, a simulation-guided framework for
action-conditioned robot video prediction.
A simulation world model serves as a training-time teacher,
transferring motion knowledge to a real-domain student through
latent-motion distillation.
Simulation-conditioning classifier-free guidance complements these internalized
priors with controllable guidance from predicted simulation
latents, requiring only the final student model at inference.
SimForcing achieves the best prediction fidelity and perceptual
similarity among the compared methods without embodied
pretraining, while downstream experiments provide preliminary
evidence of its utility for policy learning.

\newpage
\appendix
\setcounter{table}{0}
\setcounter{figure}{0}
\renewcommand{\thetable}{A\arabic{table}}
\renewcommand{\thefigure}{A\arabic{figure}}

\appendix
\section{Appendix}
\label{app:in}

\begin{figure*}[t]
    \centering
    \includegraphics[width=\textwidth]{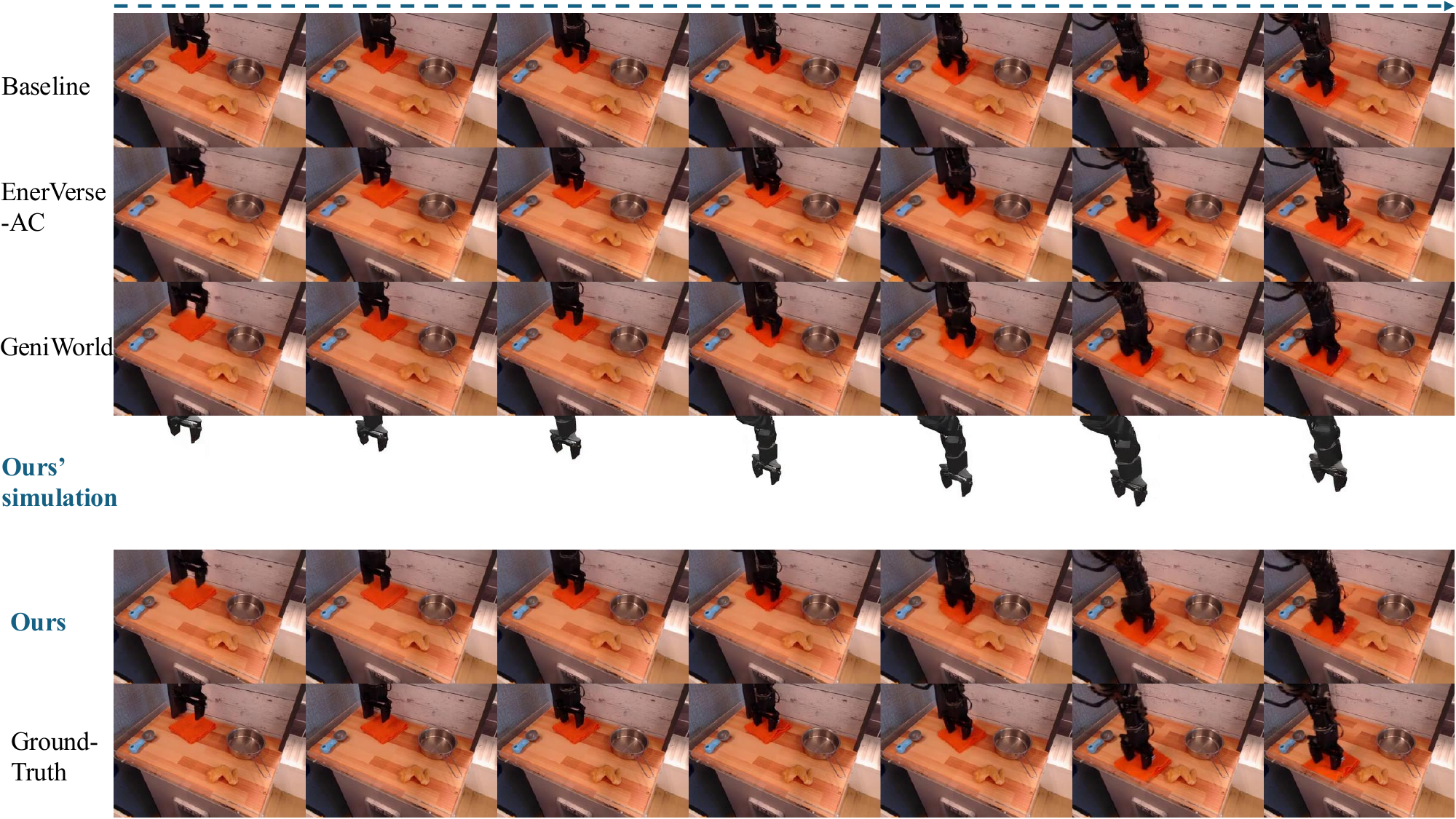}
    \caption{
        Qualitative comparison on Bridge: example 1. From top to bottom: Baseline~\citep{wan2025wan}, EnerVerse-AC~\citep{jiang2025enerverse}, GeniWorld~\citep{gu2026geniworld},
        our simulator-domain predictions, our real-world
        predictions, and ground truth. The seven columns show aligned temporal snapshots.
    }
    \label{fig:bridge_case1}
\end{figure*}

\begin{figure*}[t]
    \centering
    \includegraphics[width=\textwidth]{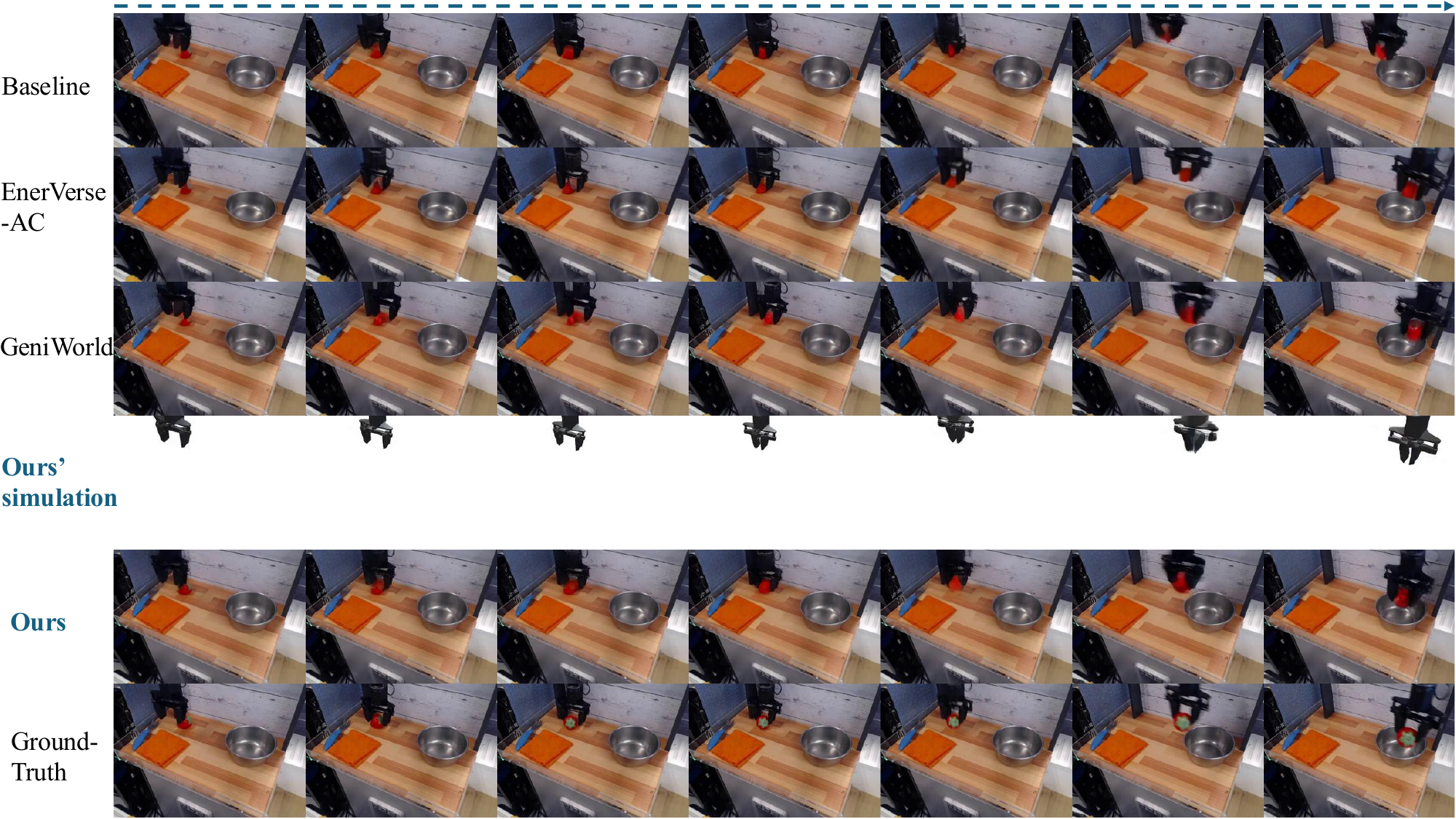}
    \caption{
        Qualitative comparison on Bridge: example 2.
        Row order and temporal sampling are identical
        to Figure~\ref{fig:bridge_case1}.
    }
    \label{fig:bridge_case2}
\end{figure*}

\subsection{Additional Qualitative Results}
\label{app:qualitative_results}

We present additional qualitative comparisons on Bridge~\citep{walke2023bridgedata} and
InternData-A1~\citep{tian2026interndata}, with four examples from each dataset.

Each example contains seven uniformly spaced
frames from the temporal interval shared by all methods.
The comparisons allow inspection of object appearance,
robot motion, and agreement with the ground-truth sequence.
We additionally show the simulator-domain predictions produced
by our method alongside its real-world predictions.

Across all figures, rows from top to bottom show
Baseline~\citep{wan2025wan}, EnerVerse-AC~\citep{jiang2025enerverse}, GeniWorld~\citep{jiang2025enerverse}, our simulator-domain
predictions, our real-world predictions, and ground truth.
Our simulator-domain and real-world predictions use the same
temporal indices.

\subsubsection{Qualitative Results on Bridge}
\label{app:qualitative_bridge}

\begin{figure*}[t]
    \centering
    \includegraphics[width=\textwidth]{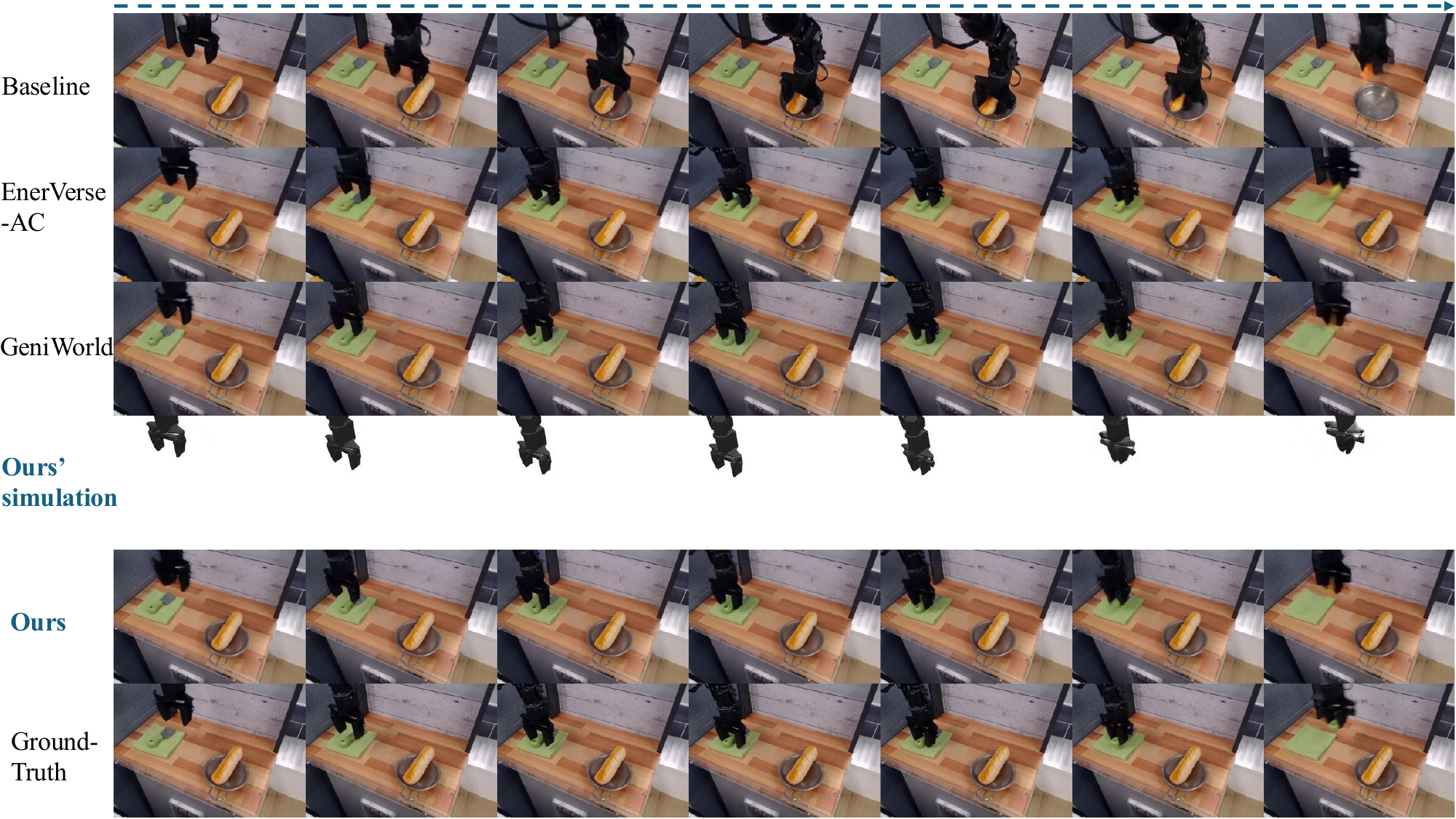}
    \caption{
        Qualitative comparison on Bridge: example 3.
        Row order and temporal sampling are identical
        to Figure~\ref{fig:bridge_case1}.
    }
    \label{fig:bridge_case3}
\end{figure*}

Figures~\ref{fig:bridge_case1}--\ref{fig:bridge_case3}
present four qualitative examples on Bridge~\citep{walke2023bridgedata}.
Columns correspond to temporally aligned frames.

\subsubsection{Qualitative Results on InternData-A1}
\label{app:qualitative_interndata_a1}

\begin{figure*}[t]
    \centering
    \includegraphics[width=\textwidth]{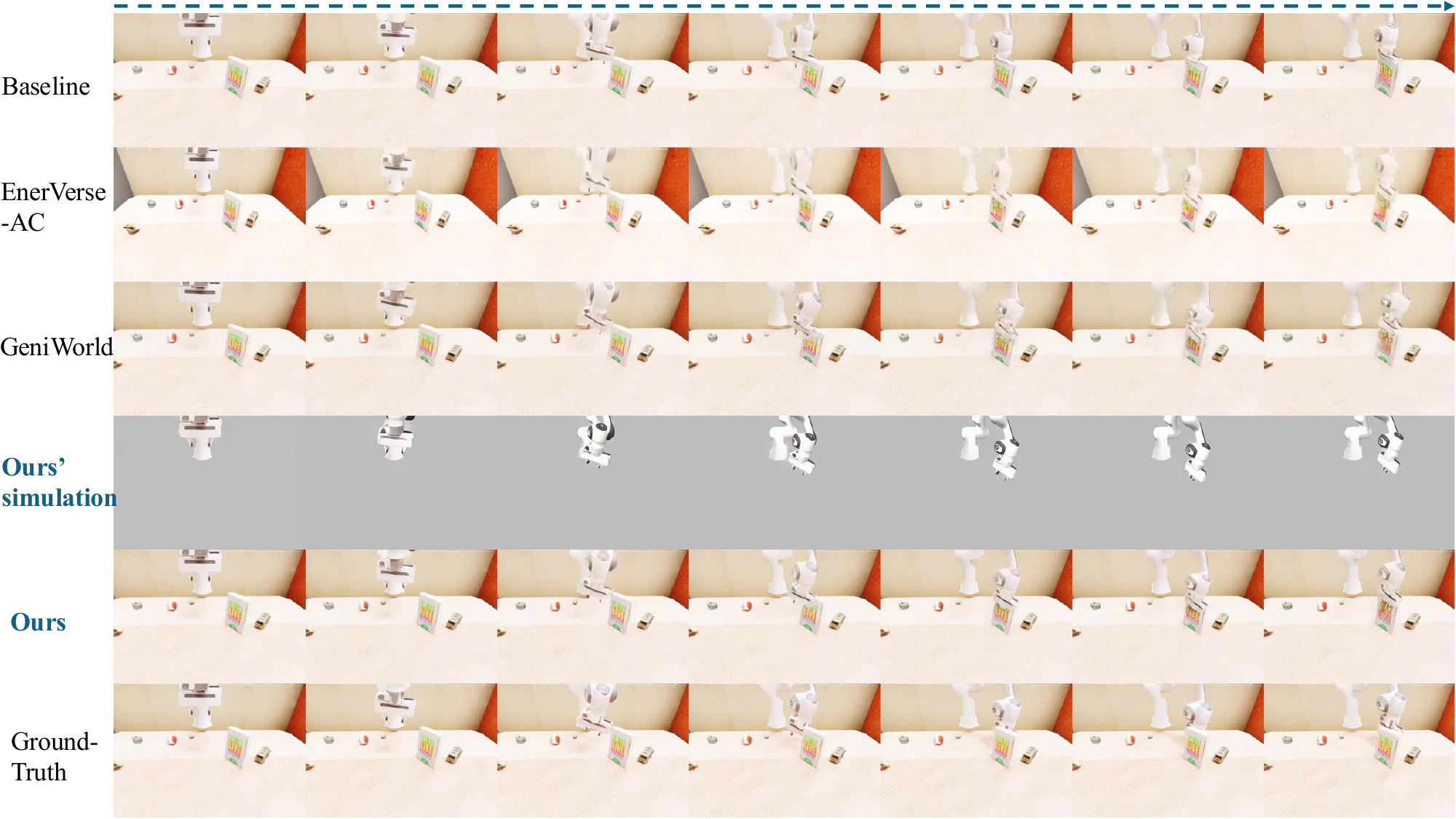}
    \caption{
        Qualitative comparison on InternData-A1: example 1.
        From top to bottom: Baseline~\citep{wan2025wan}, EnerVerse-AC~\citep{jiang2025enerverse}, GeniWorld~\citep{gu2026geniworld},
        our simulator-domain predictions, our real-world
        predictions, and ground truth. The seven columns show aligned temporal snapshots.
    }
    \label{fig:interndata_a1_case1}
\end{figure*}

\begin{figure*}[h]
    \centering
    \includegraphics[width=\textwidth]{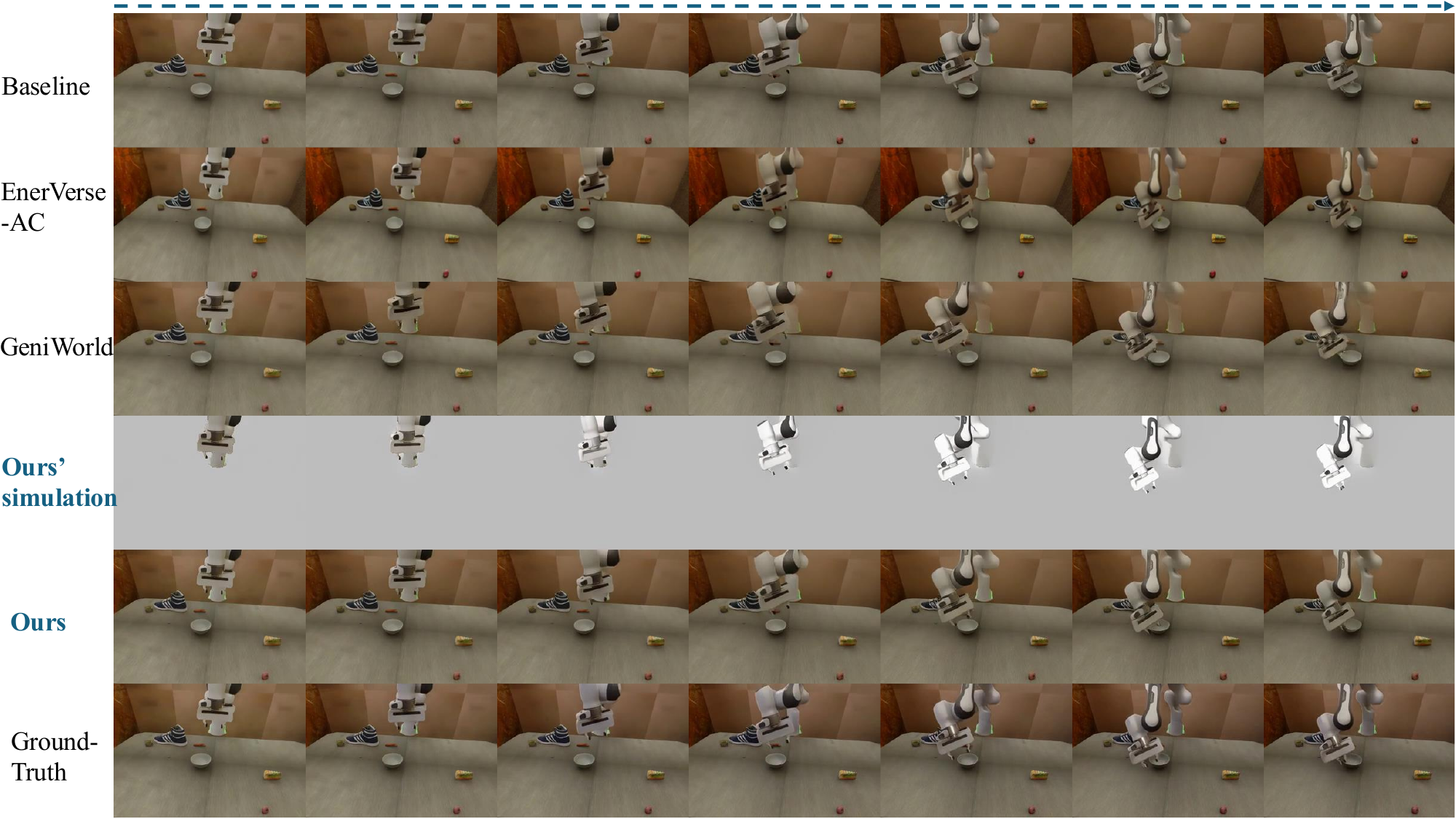}
    \caption{
        Qualitative comparison on InternData-A1: example 2.
        Row order and temporal sampling are identical
        to Figure~\ref{fig:interndata_a1_case1}.
    }
    \label{fig:interndata_a1_case2}
\end{figure*}

\begin{figure*}[h]
    \centering
    \includegraphics[width=\textwidth]{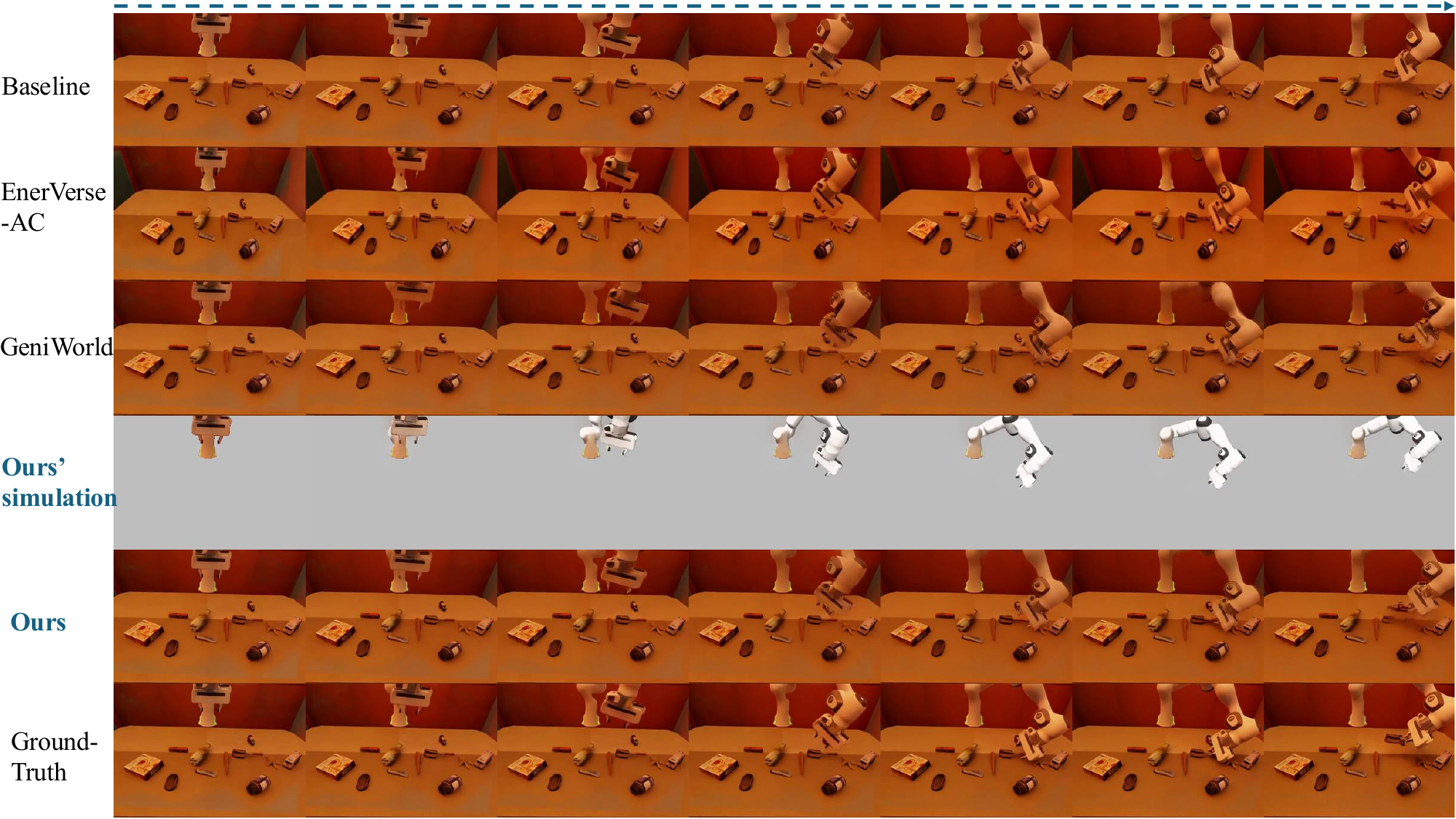}
    \caption{
        Qualitative comparison on InternData-A1: example 3.
        Row order and temporal sampling are identical
        to Figure~\ref{fig:interndata_a1_case1}.
    }
    \label{fig:interndata_a1_case3}
\end{figure*}

Figures~\ref{fig:interndata_a1_case1}--\ref{fig:interndata_a1_case3}
present four qualitative examples on InternData-A1~\citep{tian2026interndata}.
Columns correspond to temporally aligned frames.

\subsection{Effect of Training-Time Conditioning Probability}
\label{app:conditioning_probability}

\begin{table*}[t]
    \centering
    \small
    \setlength{\tabcolsep}{5pt}
    \caption{Joint evaluation of training-time conditioning probability $p$
    and inference-time guidance weight $w$. Higher PSNR (dB) and SSIM are
    better. Bold indicates the best result in the entire table for each
    metric.}
    \label{tab:conditioning_probability}
    \begin{tabular}{c cc cc cc cc}
        \toprule
        & \multicolumn{2}{c}{$w=0$}
        & \multicolumn{2}{c}{$w=0.3$}
        & \multicolumn{2}{c}{$w=0.6$}
        & \multicolumn{2}{c}{$w=1.0$} \\
        \cmidrule(lr){2-3} \cmidrule(lr){4-5}
        \cmidrule(lr){6-7} \cmidrule(lr){8-9}
        $p$ & PSNR $\uparrow$ & SSIM $\uparrow$
            & PSNR $\uparrow$ & SSIM $\uparrow$
            & PSNR $\uparrow$ & SSIM $\uparrow$
            & PSNR $\uparrow$ & SSIM $\uparrow$ \\
        \midrule
        0.0 & 24.441 & 0.8524 & 24.441 & 0.8524 & 24.441 & 0.8524 & 24.441 & 0.8524 \\
        0.3 & 24.675 & 0.8547 & 24.825 & 0.8561 & 24.890 & 0.8567 & 24.896 & 0.8566 \\
        0.7 & 24.620 & 0.8549 & 25.014 & 0.8578 & \textbf{25.077} & \textbf{0.8583} & 24.857 & 0.8567 \\
        1.0 & 21.832 & 0.8408 & 23.570 & 0.8502 & 24.387 & 0.8536 & 24.009 & 0.8508 \\
        \bottomrule
    \end{tabular}
\end{table*}

\paragraph{Experimental setup.}
We evaluate multi-block dropout conditioning on the Bridge
validation set by varying the training-time condition retention
probability $p\in\{0,0.3,0.7,1.0\}$ and inference-time CFG weight
$w\in\{0,0.3,0.6,1.0\}$.
Here, $p$ is the probability of enabling simulation conditioning
for each training sample, with the same gate shared across all
video transformer blocks.
Each checkpoint generates its own simulation conditions and
real-video predictions, so the comparison evaluates the complete
pipeline. We report PSNR and SSIM on the same evaluation samples.
The $p=0$ checkpoint uses fewer training steps than the others.

\paragraph{Effect of multi-block dropout conditioning.}
Table~\ref{tab:conditioning_probability} shows that intermediate
condition retention probabilities achieve higher best-observed
PSNR and SSIM than $p=0$ or $p=1$ in this sweep.
For $p=0$, both metrics remain unchanged across guidance weights.
For $p=1$, removing simulation conditioning at inference
substantially reduces prediction quality, suggesting strong
dependence on the condition.
In contrast, $p=0.3$ and $p=0.7$ maintain better performance
without conditioning while also benefiting from simulation guidance.
These results support the dropout mechanism in multi-block
dropout conditioning, which trains the student to predict
with and without simulation inputs.

\paragraph{Interaction with CFG.}
The configuration $(p,w)=(0.7,0.6)$ achieves the highest PSNR
and SSIM among the evaluated settings, supporting our default
choice.
For $p=0.7$, $w=0.6$ outperforms both $w=0$ and $w=1$,
indicating that moderate simulation guidance improves prediction
over either branch alone.
For $p=0.3$, PSNR peaks at $w=1$ and SSIM at $w=0.6$,
although the differences between these settings are small.
These results show that the benefit of CFG depends on
the condition retention probability used during training.

\subsection{Inference Steps and Simulation Guidance}
\label{app:inference_steps}

\paragraph{Evaluation protocol.}
We investigate the quality--latency trade-off by varying the simulation
generation steps $N_s$, real-video generation steps $N_r$, and
simulation-conditioning CFG weight $w$.
We evaluate all combinations of $N_s,N_r\in\{1,5,10,20\}$ and
$w\in\{0,0.3,0.6,1\}$, yielding 64 configurations on the same 100 Bridge
validation samples and initial observations.
Both generation stages use the same student checkpoint, with masked
initial observations as simulation inputs.
PSNR and SSIM are averaged over all 100 samples.
Latency is averaged over the remaining 99 samples after excluding the
first warm-up sample for each configuration. It includes simulation
generation, real-video generation, and real-video decoding, but excludes
model loading, metric computation, and file saving.
Although simulation predictions are shared across the four CFG weights
during evaluation, each reported latency includes the full simulation
generation cost for one configuration.

\paragraph{Effect of inference steps.}
Tables~\ref{tab:steps_psnr} and~\ref{tab:steps_ssim} show that increasing
real-video generation from one to five or ten steps substantially improves
prediction fidelity, while increasing from ten to twenty steps provides
only small additional gains.
Increasing simulation steps has a much smaller effect and does not
consistently improve either metric.
The highest observed PSNR and SSIM occur at $(N_s,N_r,w)=(5,20,0.6)$,
reaching 25.090~dB and 0.8585, respectively.
At $w=0.6$, even one simulation step produces results close to those
obtained with larger simulation budgets, suggesting that a coarse
predicted simulation reference can already provide useful guidance.

\paragraph{Guidance and latency.}
The weight $w=0.6$ yields the highest PSNR and SSIM for every tested
step combination. At $w=0$, both metrics are invariant to $N_s$,
consistent with the absence of simulation conditioning in the real-video
prediction branch.
Table~\ref{tab:steps_latency} shows that $(1,10,0.6)$ achieves
25.063~dB PSNR and 0.8575 SSIM in 2.008~s per sample: its PSNR is
0.027~dB below the highest observed value, with approximately 47\%
lower latency than $(5,20,0.6)$.
For a stronger emphasis on speed, $(1,10,1)$ reduces latency to
1.245~s, with 24.838~dB PSNR and 0.8559 SSIM.
These results indicate that inference cost can be reduced by allocating
fewer steps to simulation generation and avoiding excessive real-video
sampling steps.

\paragraph{Timing interpretation and limitations.}
In the evaluated implementation, $w=1$ uses a single forward pass per
real-video sampling step, whereas the other weights use the two-branch
CFG path. The recorded $w=0$ latency also includes simulation generation;
it therefore does not represent an optimized configuration that omits
unused simulation predictions and the unused conditional branch.
The $N_s=20$ configurations were evaluated in a separate run, so their
timings may be affected by differences in machine load.
Results are from a single sweep without repeated-run uncertainty
estimates; small metric differences should not be interpreted as
statistically established improvements.

\begin{table}[]
    \centering
    \small
    \setlength{\belowcaptionskip}{6pt}
    \caption{PSNR (dB) across inference-step budgets and CFG weights on the 100-sample Bridge evaluation. $N_s$ and $N_r$ denote simulation and real-video generation steps. Bold indicates the best result within each row.}
    \label{tab:steps_psnr}
    \begin{tabular*}{0.94\linewidth}{@{\extracolsep{\fill}}cccccc@{}}
        \toprule
        \multicolumn{2}{c}{Inference steps} & \multicolumn{4}{c}{CFG weight $w$} \\
        \cmidrule(lr){1-2}\cmidrule(lr){3-6}
        $N_s$ & $N_r$ & $0$ & $0.3$ & $0.6$ & $1$ \\
        \midrule
        1 & 1 & 22.785 & 23.116 & \textbf{23.301} & 23.206 \\
        1 & 5 & 24.263 & 24.697 & \textbf{24.851} & 24.629 \\
        1 & 10 & 24.576 & 24.958 & \textbf{25.063} & 24.838 \\
        1 & 20 & 24.620 & 24.983 & \textbf{25.063} & 24.843 \\
        \addlinespace[3pt]
        5 & 1 & 22.785 & 23.130 & \textbf{23.314} & 23.177 \\
        5 & 5 & 24.263 & 24.732 & \textbf{24.847} & 24.587 \\
        5 & 10 & 24.576 & 24.995 & \textbf{25.083} & 24.847 \\
        5 & 20 & 24.620 & 25.019 & \textbf{25.090} & 24.872 \\
        \addlinespace[3pt]
        10 & 1 & 22.785 & 23.127 & \textbf{23.307} & 23.167 \\
        10 & 5 & 24.263 & 24.727 & \textbf{24.836} & 24.574 \\
        10 & 10 & 24.576 & 24.990 & \textbf{25.070} & 24.832 \\
        10 & 20 & 24.620 & 25.014 & \textbf{25.077} & 24.857 \\
        \addlinespace[3pt]
        20 & 1 & 22.785 & 23.125 & \textbf{23.304} & 23.163 \\
        20 & 5 & 24.263 & 24.725 & \textbf{24.829} & 24.566 \\
        20 & 10 & 24.576 & 24.989 & \textbf{25.062} & 24.823 \\
        20 & 20 & 24.620 & 25.012 & \textbf{25.070} & 24.847 \\
        \bottomrule
    \end{tabular*}
\end{table}

\begin{table}[]
    \centering
    \small
    \setlength{\belowcaptionskip}{6pt}
    \caption{SSIM across inference-step budgets and CFG weights on the 100-sample Bridge evaluation. $N_s$ and $N_r$ denote simulation and real-video generation steps. Bold indicates the best result within each row.}
    \label{tab:steps_ssim}
    \begin{tabular*}{0.94\linewidth}{@{\extracolsep{\fill}}cccccc@{}}
        \toprule
        \multicolumn{2}{c}{Inference steps} & \multicolumn{4}{c}{CFG weight $w$} \\
        \cmidrule(lr){1-2}\cmidrule(lr){3-6}
        $N_s$ & $N_r$ & $0$ & $0.3$ & $0.6$ & $1$ \\
        \midrule
        1 & 1 & 0.8231 & 0.8264 & \textbf{0.8280} & 0.8271 \\
        1 & 5 & 0.8483 & 0.8518 & \textbf{0.8528} & 0.8510 \\
        1 & 10 & 0.8539 & 0.8569 & \textbf{0.8575} & 0.8559 \\
        1 & 20 & 0.8549 & 0.8576 & \textbf{0.8582} & 0.8567 \\
        \addlinespace[3pt]
        5 & 1 & 0.8231 & 0.8266 & \textbf{0.8281} & 0.8268 \\
        5 & 5 & 0.8483 & 0.8520 & \textbf{0.8527} & 0.8504 \\
        5 & 10 & 0.8539 & 0.8571 & \textbf{0.8578} & 0.8559 \\
        5 & 20 & 0.8549 & 0.8579 & \textbf{0.8585} & 0.8569 \\
        \addlinespace[3pt]
        10 & 1 & 0.8231 & 0.8266 & \textbf{0.8281} & 0.8267 \\
        10 & 5 & 0.8483 & 0.8520 & \textbf{0.8526} & 0.8502 \\
        10 & 10 & 0.8539 & 0.8571 & \textbf{0.8576} & 0.8557 \\
        10 & 20 & 0.8549 & 0.8578 & \textbf{0.8583} & 0.8567 \\
        \addlinespace[3pt]
        20 & 1 & 0.8231 & 0.8266 & \textbf{0.8281} & 0.8266 \\
        20 & 5 & 0.8483 & 0.8520 & \textbf{0.8526} & 0.8502 \\
        20 & 10 & 0.8539 & 0.8571 & \textbf{0.8576} & 0.8556 \\
        20 & 20 & 0.8549 & 0.8578 & \textbf{0.8583} & 0.8566 \\
        \bottomrule
    \end{tabular*}
\end{table}

\begin{table}[]
    \centering
    \small
    \setlength{\belowcaptionskip}{6pt}
    \caption{Mean inference latency (s/sample) across inference-step budgets and CFG weights.}
    \label{tab:steps_latency}
    \begin{tabular*}{0.94\linewidth}{@{\extracolsep{\fill}}cccccc@{}}
        \toprule
        \multicolumn{2}{c}{Inference steps} & \multicolumn{4}{c}{CFG weight $w$} \\
        \cmidrule(lr){1-2}\cmidrule(lr){3-6}
        $N_s$ & $N_r$ & $0$ & $0.3$ & $0.6$ & $1$ \\
        \midrule
        1 & 1 & 0.644 & 0.649 & 0.649 & 0.571 \\
        1 & 5 & 1.251 & 1.258 & 1.259 & 0.879 \\
        1 & 10 & 2.002 & 2.008 & 2.008 & 1.245 \\
        1 & 20 & 3.519 & 3.546 & 3.539 & 2.012 \\
        \addlinespace[3pt]
        5 & 1 & 0.934 & 0.941 & 0.940 & 0.862 \\
        5 & 5 & 1.543 & 1.557 & 1.553 & 1.168 \\
        5 & 10 & 2.298 & 2.304 & 2.304 & 1.536 \\
        5 & 20 & 3.799 & 3.802 & 3.805 & 2.286 \\
        \addlinespace[3pt]
        10 & 1 & 1.288 & 1.293 & 1.292 & 1.215 \\
        10 & 5 & 1.890 & 1.895 & 1.895 & 1.512 \\
        10 & 10 & 2.658 & 2.664 & 2.661 & 1.899 \\
        10 & 20 & 4.175 & 4.175 & 4.179 & 2.652 \\
        \addlinespace[3pt]
        20 & 1 & 1.946 & 1.951 & 1.951 & 1.876 \\
        20 & 5 & 2.551 & 2.557 & 2.558 & 2.183 \\
        20 & 10 & 3.284 & 3.295 & 3.295 & 2.547 \\
        20 & 20 & 4.772 & 4.775 & 4.777 & 3.287 \\
        \bottomrule
    \end{tabular*}
\end{table}

\subsection{Implementation Details}
\label{app:implementation}

\paragraph{BridgeData V2.}
We use a paired simulation--real subset of BridgeData V2~\citep{walke2023bridgedata}
containing 1,952 robot manipulation trajectories.
Each pair consists of temporally aligned simulated and real-world
videos sharing the same action sequence, proprioceptive states,
and language instruction.
We retain trajectories containing at least 21 frames and randomly
sample 30 temporal windows per eligible trajectory in each epoch.
Each window contains 21 consecutive video frames and 20 aligned
actions, with one action per video transition.
Both actions and proprioceptive states are seven-dimensional.
For evaluation, we select 100 paired samples from the validation set.

\paragraph{InternData-A1.}
We additionally use the Franka manipulation subset of
InternData-A1~\citep{tian2026interndata}.
Each training example pairs a simulation rendering with its
corresponding real-camera video, sharing the action sequence,
proprioceptive states, and language instruction.
The training split contains 2,657 trajectories.
We retain paired sequences that support a complete 129-step window
and randomly sample 30 windows per eligible trajectory per epoch.
Video frames are sampled every four time steps, yielding 33 frames
per window, while all 128 actions are retained.
Each video transition therefore corresponds to four consecutive
actions.
The seven-dimensional actions comprise a six-dimensional
end-effector pose and a gripper-opening command.
The seven-dimensional proprioceptive states comprise the
end-effector pose and gripper position.
For evaluation, we select 80 paired samples from the validation set.

\paragraph{Data preprocessing.}
For both datasets, simulated and real-world videos undergo the
same aspect-ratio-preserving resizing and center cropping to
$224 \times 320$ pixels.
Pixel values are normalized to $[-1,1]$.
Actions and proprioceptive states are normalized using
dataset-level minimum and maximum statistics.
The paired videos use identical temporal windows to preserve
alignment between visual observations and actions.
Language instructions are represented by precomputed text
embeddings with a maximum sequence length of 128 tokens.

\paragraph{Model initialization and conditioning.}
Our video backbone is based on Wan2.2-TI2V-5B~\citep{wan2025wan}.
For each dataset, the student is initialized from an
action-conditioned simulation world model, and a frozen copy
of the same model serves as the motion teacher.
The video backbone contains 30 transformer blocks with a hidden
dimension of 3,072 and 24 attention heads.
The action transformer contains 30 blocks with a hidden
dimension of 1,024.

Motion supervision matches adjacent-frame differences in the
estimated clean video latents and is applied to every training
sample.
The real-world prediction branch additionally receives paired
simulation latents with probability $0.7$.
The residual projections used to inject this condition are
initialized to zero.
The simulation prediction branch does not receive this auxiliary
condition.
Initial-frame and action conditions are retained throughout
training, and the strength of motion supervision is unchanged
when the auxiliary simulation condition is present.

\paragraph{Training objective.}
We jointly optimize flow matching in both domains and
teacher-guided motion supervision:
\begin{equation}
\begin{aligned}
\mathcal{L}
={}&
\lambda_{\mathrm{FM}}^{r}\mathcal{L}_{\mathrm{FM}}^{r}
+
\lambda_{\mathrm{FM}}^{s}\mathcal{L}_{\mathrm{FM}}^{s}
\\
&+
\lambda_{\mathrm{motion}}^{s2r}
\mathcal{L}_{\mathrm{motion}}^{s2r}
+
\lambda_{\mathrm{motion}}^{s2s}
\mathcal{L}_{\mathrm{motion}}^{s2s}.
\end{aligned}
\end{equation}
The loss weights for the two datasets are summarized in
Table~\ref{tab:implementation_settings}.
No action-prediction loss is used.

\begin{table}[t]
\centering
\caption{Dataset-specific training settings. Batch sizes count
paired simulation--real samples per device.}
\label{tab:implementation_settings}
\begin{tabular}{lcc}
\hline
Setting & BridgeData V2 & InternData-A1 \\
\hline
Video frames per window & 21 & 33 \\
Actions per window & 20 & 128 \\
Actions per video transition & 1 & 4 \\
Batch size per device & 16 & 10 \\
$\lambda_{\mathrm{FM}}^{r}$ & 1.0 & 1.0 \\
$\lambda_{\mathrm{FM}}^{s}$ & 0.5 & 0.5 \\
$\lambda_{\mathrm{motion}}^{s2r}$ & 1.0 & 1.0 \\
$\lambda_{\mathrm{motion}}^{s2s}$ & 0.1 & 1.0 \\
\hline
\end{tabular}
\end{table}

\paragraph{Optimization.}
For both datasets, we train for 20 epochs using AdamW with a
learning rate of $10^{-4}$, momentum coefficients
$(\beta_1,\beta_2)=(0.9,0.95)$, and weight decay $10^{-2}$.
The learning rate is linearly warmed up over the first $5\%$
of training steps and subsequently decayed with a cosine schedule.
We perform one optimizer update per batch without gradient
accumulation.
Training uses bfloat16 mixed precision, gradient checkpointing,
and gradient clipping with a maximum norm of $1.0$.
The video autoencoder and motion teacher remain frozen
throughout training.
We use 1,000 diffusion training timesteps with a timestep-shift
factor of $5.0$ and set the random seed to 42.

\paragraph{Baseline implementations.}
For IRASim~\citep{zhu2024irasim} and
Cosmos-Predict2.5~\citep{ali2025cosmos2.5}, we directly evaluate
the released checkpoints trained on Bridge, without additional
fine-tuning.
For Ctrl-World~\citep{guo2025ctrlworld} and
EnerVerse-AC~\citep{jiang2025enerverse}, we use their publicly
available code to train on our training splits for a number
of optimization steps comparable to SimForcing.
As no public implementation of GeniWorld~\citep{gu2026geniworld}
was available when we conducted the experiments, we implement
its approach based on the paper using
Wan2.2-TI2V-5B~\citep{wan2025wan}.
Our GeniWorld implementation shares the same backbone
architecture as SimForcing to facilitate a controlled comparison.

\subsection{Downstream Policy Learning on LIBERO}
\label{app:libero}

\paragraph{Experimental setup.}
We evaluate the transferability of our pretrained world model to
robotic manipulation on the four LIBERO suites: Spatial, Object,
Goal, and Long.
As summarized in Table~\ref{tab:libero}, we compare Wan2.2-based
initialization with our world-model initialization under three
downstream training settings.
The policy uses a video expert and an action expert coupled through
attention.
Its observations comprise an external-camera image and a wrist-camera
image, concatenated spatially, together with the task instruction
and the current proprioceptive state.
The action expert predicts a sequence of end-effector motion and
gripper commands.
For the Wan2.2 baseline, the video expert is initialized from Wan2.2
and the action expert from the corresponding Wan2.2-derived
ActionDiT initialization.
For Our WM, both experts inherit the compatible weights of our
pretrained world-model checkpoint.
Thus, this comparison evaluates the transfer of the pretrained
two-expert model, rather than a replacement of the visual backbone alone.

The following three training settings follow FastWAM~\citep{yuan2026fastwam};
please refer to the original paper for further details.

\paragraph{Action-only training.}
In this setting, the video expert is frozen and serves as a visual
feature extractor for the current observation.
Its layer-wise keys and values provide visual context for the action
expert, which learns to denoise action trajectories conditioned on
this context, the task instruction, and proprioception.
Only the action expert and the proprioceptive projection are optimized.
Future video frames are not used as prediction targets, and no
video-generation loss is applied.
This setting tests whether the pretrained model provides a useful
initialization for policy learning when downstream adaptation is
restricted to the action pathway.

\paragraph{Unconditional training.}
Here, both experts are optimized using video and action flow-matching
objectives.
Future video latents and action trajectories are independently
corrupted with noise, while the initial observation remains clean
and provides visual conditioning.
The video expert predicts future visual dynamics without receiving
action tokens or ground-truth actions.
The action expert attends to its own action tokens and the video
tokens of the initial observation, but cannot attend to future
video tokens.
The initial observation tokens are also prevented from attending
to future frames, avoiding indirect access to future observations
through the visual context.
Consequently, video prediction supplies an auxiliary learning
objective for the visual representation, while action prediction
depends only on information available at decision time.
The term \emph{unconditional} refers specifically to the absence of
action conditioning in the video branch; visual, language, and
proprioceptive conditioning are retained.
At inference, actions can be generated from the current observation
without generating a future video.

\paragraph{Video + action training.}
This setting retains the video and action flow-matching objectives
and jointly optimizes both experts, but expands the action expert's
attention to the complete video token sequence.
Action prediction can therefore use both the current observation
and the evolving representations of future video frames.
The coupling is directional: action tokens attend to video tokens,
whereas video tokens do not attend to action tokens.
During training, the future video tokens are noisy versions of
demonstration latents.
During inference, future video latents and actions are generated
together from noise, with the initial observation fixed.
Thus, the additional visual context available to the policy comes
from predicted futures rather than ground-truth future observations.

\paragraph{Results and interpretation.}
Our world-model initialization improves the reported average success
rate in all three settings: from $82.6\%$ to $89.2\%$ under action-only
training, from $96.5\%$ to $97.5\%$ under unconditional training, and
from $98.1\%$ to $98.4\%$ under video + action training.
These correspond to gains of $6.6$, $1.0$, and $0.3$ percentage points,
respectively.
The largest improvement occurs under action-only adaptation,
including gains of $9.6$ points on Goal and $15.2$ points on Long.
This result supports the usefulness of the transferred initialization
even when the visual expert is frozen.
The gains are smaller when both experts receive downstream video
and action supervision.
Video + action training with Our WM achieves the highest reported
average success rate, $98.4\%$, compared with $97.6\%$ for the FastWAM
reference.
The improvements are not uniform across suites: in the video + action
setting, gains on Object and Long offset decreases on Spatial and Goal.
Overall, the results support improved downstream transfer across the
three training settings, without implying an improvement on every
individual suite.



\setlength{\bibhang}{0pt}
\setlength{\bibindent}{0pt}
\bibliography{references}
\bibliographystyle{plainnat}

\end{document}